\documentclass[11pt]{article}
\usepackage[margin=1.1in]{geometry}
\usepackage[utf8]{inputenc}
\usepackage[T1]{fontenc}
\usepackage{lmodern}
\usepackage{microtype}
\usepackage{amsmath,amssymb,amsfonts,amsthm}
\usepackage{mathtools}
\usepackage{booktabs}
\usepackage{graphicx}
\usepackage[numbers]{natbib}
\usepackage{hyperref}
\usepackage{cleveref}
\hypersetup{pdftitle={Attention as Frustrated Synchronization},pdfauthor={Joshua Nunley}}

\newcommand{\sfnThirty}{1.6050}
\newcommand{\sfnThirtyStd}{0.0038}
\newcommand{\sfnConverged}{1.5953}
\newcommand{\sfnConvergedStd}{0.0014}
\newcommand{\nophaseThirty}{1.6031}
\newcommand{\nophaseThirtyStd}{0.0018}
\newcommand{\mfThirty}{1.6452}
\newcommand{\mfThirtyStd}{0.0070}
\newcommand{\mfwThirty}{1.6405}
\newcommand{\mfwThirtyStd}{0.0030}
\newcommand{\xfThirty}{1.6258}
\newcommand{\xfThirtyStd}{0.0019}
\newcommand{\xfConverged}{1.611}

\newcommand{\tanConverged}{1.6272}           % converged 50-ep Kuramoto-attention baseline (val)
\newcommand{\sfnCrossEpochs}{19, 23, 24, 19, and 27}

\newcommand{\leanEpEleven}{1.6300}

\newcommand{\sfnCodeThirty}{1.2235}
\newcommand{\sfnCodeBest}{1.2206}
\newcommand{\sfnCodeSOne}{1.3982}
\newcommand{\sfnCodeSTwo}{1.2254}
\newcommand{\xfCodeThirty}{1.2409}
\newcommand{\xfCodeThirtyStd}{0.0075}
\newcommand{\nophaseCodeVal}{1.2217}

\newcommand{\sfnCodeFiveTwelve}{1.0824}

\newcommand{\xfCodeFiveTwelve}{1.0931}
\newcommand{\codeMarginFiveTwelve}{-0.011}
\newcommand{\codeMarginTwoFiftySix}{-0.025}

\newcommand{\ladderTwoFSN}{1.5264}
\newcommand{\ladderTwoXFFinal}{1.5521}
\newcommand{\ladderFourFSN}{1.4600}
\newcommand{\ladderFourXFFinal}{1.5060}
\newcommand{\ladderEightFSN}{1.4217}\newcommand{\ladderEightFSNEpoch}{14}
\newcommand{\ladderEightXF}{1.4497}\newcommand{\ladderEightXFFinal}{1.4368}

\newcommand{\wallCodeRatioA}{0.98}
\newcommand{\wallCodeRatioB}{1.02}
\newcommand{\wallCodeFiveTwelveRatio}{1.82}

\newcommand{\tpsFSNOneM}{263}
\newcommand{\tpsXFOneM}{965}
\newcommand{\tpsNophaseOneM}{300}
\newcommand{\tpsMFOneM}{234}
\newcommand{\tpsFSNCode}{264}
\newcommand{\tpsXFCode}{958}
\newcommand{\tpsFSNFourM}{126}
\newcommand{\tpsXFFourM}{377}
\newcommand{\memFSNOneM}{3.0}
\newcommand{\memXFOneM}{0.9}
\newcommand{\tpsEvalFSN}{1.0}
\newcommand{\tpsEvalXF}{2.7}
\newcommand{\kernelShare}{12}

\newcommand{\copyFourSeven}{-0.053}          % copy-depth margins, lean_long ep11 vs comp_xf_1m_s0 converged
\newcommand{\copyEightFifteen}{-0.094}
\newcommand{\copySixteen}{-0.133}
\newcommand{\torusSwigluPenalty}{+0.018}     % torus_swiglu_enwiki ep2 1.7081 vs nophase30_s0 ep2 1.6900 (state-dict diff = FFN only;
\newcommand{\rmsRamp}{33}                    % swiglu-std init pre-bound FFN RMS layer0->layer3: 0.000103->0.003394 = 33.0x
\newcommand{\ablNoSuccDelta}{+0.076}
\newcommand{\bndNoneEpOne}{10.19}            % bnd_none (bound removed entirely), log 7368546 ep1 val BPC 10.1926; screen protocol
\newcommand{\sfnParams}{1{,}011{,}834}       % sakanoise50/nophase30 (enwik8)
\newcommand{\xfParams}{974{,}845}            % comp_xf_1m (enwik8); FSN is 3.8% heavier on text
\newcommand{\sfnCodeParams}{973{,}818}       % sakacode30
\newcommand{\xfCodeParams}{1{,}012{,}185}    % xfmcode30; FSN is 3.8% lighter on code
\newcommand{\paramImbalance}{3.8}              % \sfnParams/\xfParams - 1 = 3.79% (comment at \xfParams)
\newcommand{\sfnEpTwoMean}{1.6892}             % sakanoise50_s0-2 ep2 seed mean (Sec 4 phase commentary)
\newcommand{\nophaseEpTwoMean}{1.6934}         % nophase30_s0-2 ep2 seed mean
\newcommand{\mfGapThirty}{0.019}               % \mfThirty - \xfThirty = 0.0194
\newcommand{\mfwGapThirty}{0.015}              % \mfwThirty - \xfThirty = 0.0147
\newcommand{\sfnMfGap}{0.040}                  % \mfThirty - \sfnThirty = 0.0402
\newcommand{\sfnMfwGap}{0.036}                 % \mfwThirty - \sfnThirty = 0.0355
\newcommand{\codeContextDropFSN}{0.141}        % ep30 256->512 improvement, FSN (comment at \codeMarginFiveTwelve)
\newcommand{\codeContextDropXF}{0.156}         % ep30 256->512 improvement, XF
\newcommand{\codeMarginFiveTwelveMean}{-0.016} % epoch-mean margin ep1-30, seq 512 (comment above)
\newcommand{\codeMarginTwoFiftySixMean}{-0.020}% epoch-mean margin ep1-30, seq 256
\newcommand{\ladderTwoLighter}{2.0}            % FSN params lighter than XF at 2M (comment at \ladderTwoFSN)
\newcommand{\ladderFourLighter}{4.0}           % 4M rung (comment at \ladderFourFSN)
\newcommand{\ladderEightLighter}{0.15}         % 8M rung: 1 - 8,073,299/8,085,645 = 0.15%
\newcommand{\wallCodeFiveTwelveSlowdown}{3.9}  % per-epoch slowdown, code seq 512 (comment above: 3.91x)
\newcommand{\slowdownOneM}{3.7}                % App B throughput ratio 965/263 = 3.67
\newcommand{\slowdownFourM}{3.0}               % App B throughput ratio 377/126 = 2.99
\newcommand{\slowdownNophaseOneM}{3.2}         % App B throughput ratio 965/300 = 3.22
\newcommand{\copySixteenStd}{-0.122}           % 16-23 bin, standard slice (comment at \copyFourSeven)
\newcommand{\copySixteenEnr}{-0.135}           % 16-23 bin, enriched slice
\newcommand{\copyTwentyFour}{-0.036}           % 24-32 bin, pooled std+enr -0.0363 (recomputed 2026-06-12
\newcommand{\copyTwentyFourStd}{-0.080}        % 24-32 bin, standard slice -0.0797
\newcommand{\copyTwentyFourEnr}{-0.035}        % 24-32 bin, enriched slice -0.0348

\newcommand{\cdNophaseEpoch}{30}               % nophase30_s0 best.pt
\newcommand{\cdXfEpoch}{48}                    % results/tmlr_runs/comp_xf_1m_s0 best.pt (converged reference)
\newcommand{\cdNtokZeroOne}{208{,}035}
\newcommand{\cdNtokTwoThree}{30{,}832}
\newcommand{\cdNtokFourSeven}{15{,}979}
\newcommand{\cdNtokEightFifteen}{5{,}019}
\newcommand{\cdNtokSixteen}{7{,}479}
\newcommand{\cdNtokTwentyFour}{13{,}058}
\newcommand{\cdBaseZeroOne}{+0.0092}
\newcommand{\cdBaseTwoThree}{+0.0193}
\newcommand{\cdBaseFourSeven}{+0.0140}
\newcommand{\cdBaseEightFifteen}{+0.0279}
\newcommand{\cdBaseSixteen}{+0.4307}
\newcommand{\cdBaseTwentyFour}{+0.0303}
\newcommand{\cdSfnZeroOne}{-0.0036}
\newcommand{\cdSfnTwoThree}{-0.0076}
\newcommand{\cdSfnFourSeven}{-0.0799}
\newcommand{\cdSfnEightFifteen}{-0.1307}
\newcommand{\cdSfnSixteen}{-0.0577}
\newcommand{\cdSfnTwentyFour}{-0.1564}
\newcommand{\cdNophaseZeroOne}{-0.0033}
\newcommand{\cdNophaseTwoThree}{+0.0005}
\newcommand{\cdNophaseFourSeven}{-0.0726}
\newcommand{\cdNophaseEightFifteen}{-0.1294}
\newcommand{\cdNophaseSixteen}{-0.0807}
\newcommand{\cdNophaseTwentyFour}{-0.1457}

\newcommand{\cdBaseEpochs}{46, 49, and 49}     % comp_tan_1m_s0/s1/s2 best.pt epochs
\newcommand{\cdSfnEpochs}{48, 50, and 50}      % sakanoise50_s0/s1/s2 best.pt epochs
\newcommand{\cdBaseZeroOneMean}{+0.0119}\newcommand{\cdBaseZeroOneRange}{0.0064}
\newcommand{\cdBaseTwoThreeMean}{+0.0170}\newcommand{\cdBaseTwoThreeRange}{0.0127}
\newcommand{\cdBaseFourSevenMean}{+0.0349}\newcommand{\cdBaseFourSevenRange}{0.0883}
\newcommand{\cdBaseEightFifteenMean}{+0.0511}\newcommand{\cdBaseEightFifteenRange}{0.0850}
\newcommand{\cdBaseSixteenMean}{+0.4231}\newcommand{\cdBaseSixteenRange}{0.4664}
\newcommand{\cdBaseSixteenMin}{+0.19}\newcommand{\cdBaseSixteenMax}{+0.65}
\newcommand{\cdBaseTwentyFourMean}{+0.0772}\newcommand{\cdBaseTwentyFourRange}{0.1082}
\newcommand{\cdSfnZeroOneMean}{-0.0054}\newcommand{\cdSfnZeroOneRange}{0.0055}
\newcommand{\cdSfnTwoThreeMean}{-0.0033}\newcommand{\cdSfnTwoThreeRange}{0.0077}
\newcommand{\cdSfnFourSevenMean}{-0.0720}\newcommand{\cdSfnFourSevenRange}{0.0183}
\newcommand{\cdSfnEightFifteenMean}{-0.1225}\newcommand{\cdSfnEightFifteenRange}{0.0233}
\newcommand{\cdSfnSixteenMean}{-0.1041}\newcommand{\cdSfnSixteenRange}{0.1337}
\newcommand{\cdSfnTwentyFourMean}{-0.1422}\newcommand{\cdSfnTwentyFourRange}{0.0367}
\newcommand{\cdNophaseZeroOneMean}{-0.0005}\newcommand{\cdNophaseZeroOneRange}{0.0055}
\newcommand{\cdNophaseTwoThreeMean}{-0.0018}\newcommand{\cdNophaseTwoThreeRange}{0.0156}
\newcommand{\cdNophaseFourSevenMean}{-0.0739}\newcommand{\cdNophaseFourSevenRange}{0.0106}
\newcommand{\cdNophaseEightFifteenMean}{-0.1301}\newcommand{\cdNophaseEightFifteenRange}{0.0045}
\newcommand{\cdNophaseSixteenMean}{-0.1631}\newcommand{\cdNophaseSixteenRange}{0.1451}
\newcommand{\cdNophaseTwentyFourMean}{-0.1341}\newcommand{\cdNophaseTwentyFourRange}{0.0176}
\newcommand{\cdCurveSeventeen}{+1.28}
\newcommand{\cdCurveNineteen}{+0.70}
\newcommand{\cdCurveTwentyTwo}{+0.36}

\newcommand{\crossAfterMin}{0.006}
\newcommand{\crossAfterMax}{0.039}

\newcommand{\cdcNtokZeroOne}{162{,}634}
\newcommand{\cdcNtokTwoThree}{30{,}576}
\newcommand{\cdcNtokFourSeven}{33{,}004}
\newcommand{\cdcNtokEightFifteen}{23{,}514}
\newcommand{\cdcNtokSixteen}{11{,}025}
\newcommand{\cdcNtokTwentyFour}{23{,}420}
\newcommand{\cdcSfnZeroOne}{-0.0106}
\newcommand{\cdcSfnTwoThree}{-0.0030}
\newcommand{\cdcSfnFourSeven}{-0.0078}
\newcommand{\cdcSfnEightFifteen}{-0.0251}
\newcommand{\cdcSfnSixteen}{-0.0470}
\newcommand{\cdcSfnTwentyFour}{-0.0518}
\newcommand{\cdcFailZeroOne}{+0.2239}
\newcommand{\cdcFailSixteen}{-0.0165}
\newcommand{\cdcFailTwentyFour}{-0.0293}
\newcommand{\cdcNophaseZeroOne}{-0.0219}
\newcommand{\cdcNophaseTwoThree}{+0.0014}
\newcommand{\cdcNophaseFourSeven}{-0.0000}
\newcommand{\cdcNophaseEightFifteen}{-0.0138}
\newcommand{\cdcNophaseSixteen}{-0.0299}
\newcommand{\cdcNophaseTwentyFour}{-0.0349}
\newcommand{\cdcDeepLo}{-0.034}
\newcommand{\cdcDeepHi}{-0.052}
\newcommand{\codeDeepShare}{3.0}
\newcommand{\enwikDeepShare}{0.6}

\newcommand{\codeFailEpOne}{1.5493}
\newcommand{\codeHealthyEpOneA}{1.3421}
\newcommand{\codeHealthyEpOneB}{1.3379}
\newcommand{\codeFiveTwelveFailEpOne}{1.4100}
\newcommand{\codeFiveTwelveHealthyEpOneA}{1.1872}
\newcommand{\codeFiveTwelveHealthyEpOneB}{1.1806}
\newcommand{\codeSeedSepMin}{0.2}
\newcommand{\mfDivergeEpSixteen}{1.6564}

\newcommand{\wcSlowOne}{3.68}
\newcommand{\wcMeanOne}{1.23}\newcommand{\wcMeanOneLo}{0.99}\newcommand{\wcMeanOneHi}{1.47}
\newcommand{\wcBestOne}{1.46}\newcommand{\wcBestOneLo}{1.22}\newcommand{\wcBestOneHi}{1.58}
\newcommand{\wcSlowTwo}{3.23}
\newcommand{\wcMeanTwo}{1.56}\newcommand{\wcMeanTwoLo}{1.51}\newcommand{\wcMeanTwoHi}{1.61}
\newcommand{\wcBestTwo}{2.00}\newcommand{\wcBestTwoLo}{1.95}\newcommand{\wcBestTwoHi}{2.05}
\newcommand{\wcSlowFour}{2.98}
\newcommand{\wcMeanFour}{0.70}\newcommand{\wcMeanFourLo}{0.69}\newcommand{\wcMeanFourHi}{0.70}
\newcommand{\wcBestFour}{1.19}\newcommand{\wcBestFourLo}{1.10}\newcommand{\wcBestFourHi}{1.29}
\newcommand{\wcSlowEight}{3.02}
\newcommand{\wcMeanEight}{0.70}
\newcommand{\wcBestEight}{0.70}
\newcommand{\C}{\mathbb{C}}
\newtheorem{proposition}{Proposition}

\title{Attention as Frustrated Synchronization}

\author{Joshua Nunley\\
Cognitive Science Program\\
Luddy School of Informatics, Computing, and Engineering\\
Indiana University Bloomington\\
\texttt{joshnunl@iu.edu}}

\date{}

\begin{document}
\maketitle

\begin{abstract}
Self-attention, applied through depth, drives token representations toward consensus, the behavior
of a synchronizing system. This makes it natural to build attention directly from coupled
oscillators, synchronizing each token toward the ones it attends to. Such consensus suits retrieval
but not prediction, because agreeing with a retrieved context records where that context is rather
than what followed it. We introduce the Frustrated Synchronization Network (FSN), an oscillator-based
attention layer that replaces consensus with frustrated synchronization, pulling each token toward a
data-determined offset from the context it attends to rather than toward agreement. Coupling a token
to the successors of the tokens it attends to makes this offset a Kuramoto--Sakaguchi frustration set
by the data itself, so the attention that retrieves a context also continues it. Because every
coefficient of the resulting coupling kernel is a named object from the synchronization literature,
the trained layer can be read directly as a coupling function. At one million matched parameters on
character-level text and code, the FSN reaches lower validation loss than a tuned transformer, with
its advantage concentrated on long-range copying, and a fully oscillator-native variant approaches
the transformer's quality with no feed-forward network.
\end{abstract}

\section{Introduction}
\label{sec:intro}

Self-attention, applied through depth, drives token representations toward consensus. Across layers
their states cluster, lose rank, and collapse toward a shared
value~\citep{dong2021attention,geshkovski2023emergence,abella2025consensus}. This is the behavior of
a synchronizing system, coupled units settling into agreement.

Kuramoto attention~\citep{nunley2026kuramoto} builds an attention layer on this synchronization. Its
token states are phases, and each token is updated toward the phases of the tokens it attends to. At
matched parameters on enwik8 it is competitive with a tuned transformer but does not reach it. To
locate the gap, we decompose validation loss by copy depth, the length of the longest substring
ending at a position that also appears earlier in its context. Such repetition is the regime where
attention outperforms efficient alternatives~\citep{arora2024zoology}, and Kuramoto attention's
deficit against the transformer concentrates there.

The Frustrated Synchronization Network (FSN) changes what each token synchronizes toward. Rather than
pulling a token toward the tokens it attends to, it couples the token to the \emph{successor} of each
attended token, the token that followed it in the sequence. Coupling to a token's successor is
Kuramoto--Sakaguchi coupling to that token with a frustration angle equal to the token's own
transition $\delta_u = \theta_{u+1} - \theta_u$ (Proposition~\ref{prop:sakaguchi}), so the data's own
transitions set the frustration. Because every coefficient of the coupling kernel is a named object
from the synchronization literature, the trained layer can be read directly as a coupling function.

At one million matched parameters on character-level text and code, the FSN's validation loss is
below a tuned transformer's at matched epochs (\sfnThirty{} versus \xfThirty{} bits per character on
enwik8) and converges below it as well (\sfnConverged{} versus \xfConverged{}). On the copy-depth
decomposition it is ahead of the transformer on every bin of depth four and beyond, reversing a
Kuramoto-attention deficit of up to \cdBaseSixteenMean{} bits per character. The advantage at matched
epochs persists across a parameter-doubling ladder to eight million parameters. The FSN trains
several times more slowly per epoch, but against the mean transformer it reaches the transformer's
loss in \wcMeanFour$\times$ the wall-clock at four and eight million parameters and \wcMeanOne$\times$
and \wcMeanTwo$\times$ at one and two million. A variant that removes the feed-forward network,
replacing it with gated coupling to learned collective modes, reaches transformer range on enwik8
(\mfThirty{} versus \xfThirty{}) while falling short of the full FSN. These results are at character
level, and subword tokenization and larger models remain future work.

\section{The Frustrated Synchronization Network}
\label{sec:architecture}

An attention computation has two parts. The score map selects which earlier states are
relevant to the current token, and the value pathway specifies how those retrieved states act
on the current state. The FSN keeps the torus-valued content-addressed score map of the base
layer and replaces the value pathway's attractive synchronization with a frustrated coupling law.

\subsection{State and attention scores}

The FSN processes a sequence of $T$ tokens through $L$ residual layers. The state of token
$t$ at a given layer is a vector of phases $\theta_t \in \mathbb{T}^k$, and we write
$z_t = e^{i\theta_t} \in \mathbb{C}^k$ for the corresponding unit phasors, with all
operations on $z$ understood coordinatewise. A learned embedding assigns each vocabulary
item a point on the torus, and a readout scores each vocabulary item by the coherence between
the final state and a learned prototype phase vector for that item.

The attention scores are inherited from the base Kuramoto attention
layer~\citep{nunley2026kuramoto}. The score between a query token $t$ and a key token $u \le t$ is a gated
sum of phase coherences with a rotary drift,
\begin{equation}
\label{eq:score}
s_{tu} \;=\; \frac{1}{\tau}\sum_{c=1}^{k} g^{q}_{c}(t)\, g^{k}_{c}(u)\,
\cos\!\big(\theta_{t,c}-\theta_{u,c}+\omega_{c}\,(t-u)\big),
\qquad
A_{t\cdot} = \mathrm{softmax}(s_{t\cdot}),
\end{equation}
where the gates $g^{q}$ and $g^{k}$ are learned nonnegative functions of the token state,
$\tau$ is a learned temperature, and the per-coordinate angular rates $\omega_c$ implement
rotary position as a phase drift. The softmax is causal. The full gate
parameterization follows the base layer and is specified in
Appendix~\ref{app:config}.

\subsection{The coupling kernel}

The base layer applies Kuramoto coupling through the attention weights. The update direction
for token $t$ is $a_t = \sum_{u\le t} A_{tu}\sin(\theta_u - \theta_t)$, applied
coordinatewise, the gradient of phase agreement with the attended mixture. The FSN replaces
this with a learned complex kernel over harmonics $n = 1,\dots,N$ and a one-step delay. Let
$z_{+1}$ denote the sequence of phasors advanced by one position, so that the entry of
$z_{+1}$ at position $u$ is $z_{u+1}$. The update direction is
\begin{equation}
\label{eq:kernel}
a_t \;=\; \sum_{n=1}^{N} \operatorname{Im}\!\Big[\, \bar{z}_t^{\,n} \odot
\Big( A\,\big( w_0^{(n)} \odot z^{n} \;+\; w_1^{(n)} \odot z_{+1}^{n} \big) \Big)_t \Big],
\end{equation}
with learned coefficients $w_0^{(n)}, w_1^{(n)} \in \mathbb{C}^k$ per layer, where powers,
conjugation, and the product $\odot$ act coordinatewise and $A$ acts along the sequence. The
attended sum runs over strictly earlier positions. The diagonal self-contribution has the
closed form $\operatorname{Im}(w_0^{(n)})$ because $\bar{z}_t^n z_t^n = 1$, and is added
separately. The delay entries respect causality because position $u$ contributes $z_{u+1}$
only for $u + 1 \le t$. The update $a_t$ is a tangent vector to the torus at $\theta_t$. It
passes through a learned signed gate and a bounded normalization, again inherited from the
base layer, before being added to the state.

In polar form each learned scalar of Equation~\ref{eq:kernel} is a named object in the
synchronization literature. Writing $w_0^{(n)} = \rho_n e^{i\varphi_n}$ for one coordinate,
the $w_0$ part of the update is
\begin{equation}
\label{eq:sakaguchi-form}
\sum_{u} A_{tu}\, \rho_n \sin\!\big( n(\theta_u - \theta_t) + \varphi_n \big),
\end{equation}
attention-weighted Kuramoto--Sakaguchi--Daido coupling: the magnitudes $\rho_n$ are harmonic
gains and the angles $\varphi_n$ are static frustration angles. The $w_1$ part has the same
form with $\theta_{u+1}$ in place of $\theta_u$, and Section~\ref{sec:theory} shows it is
itself a Sakaguchi term whose frustration angle is supplied by the data rather than by a
parameter.

The kernel contains the base layer and several restricted couplings as exact special cases.
Setting $N=1$, $w_0 = 1$, $w_1 = 0$ recovers Kuramoto attention. Real-valued $w$ with
$N > 1$ gives Daido harmonic coupling, and $w_0 = 0$ couples purely to successors. The full FSN
uses $N=3$ with unconstrained complex coefficients per coordinate.

\subsection{Initialization of the frustration angles}
\label{sec:init}

The complex phases of the kernel require care at initialization because the zero-angle point
is symmetric. With all frustration angles at zero, the early loss surface is locally
symmetric in many of the angle directions, and in our experiments the angles fail to
differentiate before the magnitudes commit. Initializing them with small random values,
$\varphi \sim \mathcal{N}(0, \sigma^2)$ with $\sigma = 0.05$, breaks the symmetry. All FSN
results use this initialization, whose effect concentrates early in training
(Section~\ref{sec:ablations}).

\subsection{The feed-forward block and the FSN-MF variant}
\label{sec:ffn}

Between attention layers the FSN retains a standard SwiGLU feed-forward block acting on the
state, the only component that does not operate through phase coupling. The FSN-MF variant
replaces it with a small set of learned collective modes of the layer's own phases,
generalized order parameters in the sense of mean-field coupled-oscillator theory,
\begin{equation}
\label{eq:meanfield}
m_h \;=\; \sum_{c=1}^{k} C_{hc}\, z_c, \qquad
q \;=\; \mathrm{SiLU}\!\big(\operatorname{Re} m^{g} + b\big)\odot m^{v}, \qquad
a^{\mathrm{ffn}} \;=\; \operatorname{Im}\!\big(\bar{z} \odot D q\big),
\end{equation}
with learned complex mode matrices $C^{g}, C^{v}$, a learned projection $D$, and the number
of modes chosen so that the parameter count matches the SwiGLU block it replaces to within
a tenth of a percent. This removes the last non-oscillator component from the stack.

The FSN-MF reaches transformer range on enwik8, within \mfGapThirty{} bits per character of
the parameter-matched transformer baseline (Section~\ref{sec:experiments}), recovering most
of the SwiGLU block's contribution to validation loss. It does not match the full FSN, and
the gap has a specific source. The phases $\theta$ accumulate across layers without wrapping,
so the raw state encodes winding information on the universal cover $\mathbb{R}^k$ rather
than on the torus. The SwiGLU block reads this lift directly, whereas any function built from
$z = e^{i\theta}$, including Equation~\ref{eq:meanfield}, is periodic by construction and
cannot see winding. The channel the feed-forward block reads and the phasor block cannot is
thus the non-periodic lift of the phase state, and two independent measurements isolate it.
Restricting the SwiGLU's input to periodic functions of the state costs
$\torusSwigluPenalty$ bits per character at epoch two against the configuration-matched
no-phase stack, more than a third of the mean-field gap at the same checkpoint. This screen
ran the periodic-input block at roughly three times the parameters of the block it replaces,
an imbalance in the periodic block's favor that makes the penalty conservative. Separately,
the root-mean-square scale of the raw phase state the SwiGLU reads grows by a factor of
roughly $\rmsRamp$ from the first layer to the last already at initialization, the signature
of accumulated winding and absent by construction from the phasor inputs of
Equation~\ref{eq:meanfield}.

\subsection{Parameter counts and cost}

At this scale both models are sized to a one-million-parameter target by an automated
matching procedure. On enwik8 the realized counts are $\sfnParams$ for the FSN against
$\xfParams$ for the transformer, so the FSN carries \paramImbalance{} percent more
parameters on text. On the code corpus the smaller vocabulary changes the table sizes:
$\sfnCodeParams$ against $\xfCodeParams$, with the FSN the lighter of the two. The kernel
adds $2 \times N \times k$ complex coefficients, that is, $4 N k$ real parameters, per layer
over the base layer, below one percent of the total. A training step of the current FSN
implementation costs
% verifier 2026-06-12: measured throughput ratios (Appendix B) are 3.7x at 1M, 3.0x at 4M,
% 3.9x at code seq 512; the earlier "between two and three" understated the cost.
roughly three to four times as much as a training step of the transformer baseline at equal
batch size. The kernel itself adds little, and the overhead is the bounded geometric update
machinery inherited from the base layer. Appendix~\ref{app:efficiency} reports wall-clock
figures and implementation costs.
\section{Prediction as Frustrated Synchronization}
\label{sec:theory}

Consensus and continuation are distinct operations. Attention
selects the states that interact, and the coupling law determines whether those states pull the
current token toward their present phases or toward the transitions that followed them.

\subsection{Why consensus cannot predict}

The base coupling $a_t = \sum_u A_{tu} \sin(\theta_u - \theta_t)$ on a single
coordinate is the negative gradient, in $\theta_t$, of the disagreement
potential $\sum_u A_{tu}\,(1 - \cos(\theta_u - \theta_t))$, so the dynamics move every
token toward the circular mean of the tokens it attends to. At a fixed point the
configuration is phase-locked, and the readout at token $t$ sees only a summary of where the
attended context \emph{is}. Next-token prediction instead needs the state at $t$ to encode
where the context is \emph{going}, information about transitions rather than positions. No
reweighting of a pure attraction toward present phases supplies it, as the copy-depth
measurements in Section~\ref{sec:capability} confirm empirically.

\subsection{The delay term is data-dependent Sakaguchi frustration}

The Kuramoto--Sakaguchi model augments the coupling with a constant frustration angle
$\varphi$, replacing $\sin(\theta_u - \theta_t)$ by $\sin(\theta_u - \theta_t + \varphi)$,
the standard minimal model of synchronization that settles at a structured offset
rather than at agreement. The FSN's delay coupling comes from the term $w_1$ in
Equation~\ref{eq:kernel}, which couples token $t$ to the successor phase $\theta_{u+1}$ of
each attended token $u$, and this delay is exactly such a frustration.

\begin{proposition}[Delay coupling as data-dependent frustration]
\label{prop:sakaguchi}
Let $\delta_u = \theta_{u+1} - \theta_u$ denote the local transition of the attended token
$u$ on a given coordinate. Then
\begin{equation*}
\sin\!\big(\theta_{u+1} - \theta_t\big) \;=\;
\sin\!\big( (\theta_u - \theta_t) + \delta_u \big),
\end{equation*}
so coupling to the successor of $u$ is Kuramoto--Sakaguchi coupling to $u$ itself whose
frustration angle is $\delta_u$, the transition the data made at $u$.
\end{proposition}

Next-token prediction is therefore synchronization. A static Sakaguchi angle $\varphi$ makes
a network settle at a fixed offset from consensus. The delay term makes token $t$ settle at
an offset from each attended token equal to that token's own next step. When attention
selects positions whose history resembles the present, as the coherence score of
Equation~\ref{eq:score} arranges, the update moves the current state toward where analogous
contexts went next. Synchronization is frustrated by the data itself. The static phases
$\varphi_n$ and the data-dependent angles $\delta_u$ enter the coupling function in the same
algebraic slot, the former learned constants and the latter read from the local
transition at each attended token. The delay term parallels anticipating synchronization in
physics, where coupling to a time-shifted signal can lock a driven system to the future of
its driver. Here the time shift is the one-step successor and the gain is attention-weighted
and learned.

\subsection{Harmonics, signs, and transport}

The harmonic structure of the kernel has a separate role. A single sine
$\sin(\theta_u - \theta_t)$ has a bounded restoring force that vanishes as the phase gap
approaches $\pi$, so it cannot drive a token sharply onto a specific retrieved phase roughly
antipodal to it. Daido's generalization of the Kuramoto model replaces the sine by an
arbitrary odd coupling function expanded in harmonics, and the shape of that function
determines which configurations attract, and several harmonics together can sharpen the pull
toward a single target that one harmonic cannot. Trained FSNs exploit this freedom. In the
early layers they place nontrivial weight on the second and third harmonics. In the late
layers the coordinate-mean real coefficients of these harmonics flip negative, and a
negative harmonic coefficient is coupling at a phase offset of $\pi$, hence repulsion. A
coupling function with both attractive and repulsive components moves states between basins
rather than averaging them, and these late-layer sign flips drive a token's state
\emph{through} intermediate configurations toward the retrieved phase. The learned value
gates of the base layer, which multiply the coupling per coordinate, reinforce this: they
train to signed values even though the architecture permits purely positive gating, and the
sign participates in the same repulsion mechanism. The realized coupling functions are
directly readable from the trained kernel coefficients and are shown in
Section~\ref{sec:mechanism}.

The kernel thus separates three operations. The $n=1$, $w_0$ term with its gates performs
consensus: retrieval and aggregation of the attended context. The higher harmonics and the
signed structure transport states between configurations. The $w_1$ term anticipates.
Removing the delay term is the largest single subtraction in Section~\ref{sec:ablations},
consistent with anticipation as the dominant term, while removing the static phases costs
early-training speed but not converged quality.

\subsection{Generality beyond the torus}

Nothing in Equation~\ref{eq:kernel} is specific to the torus beyond the choice of group.
The state space $\mathbb{T}^k$ is the maximal torus of the unitary group, the harmonics
$z^n$ are its one-dimensional irreducible representations, and the kernel is a learned
element of the group algebra applied to the attended trajectory, with the delay term
defined by the sequence structure rather than by the geometry. On a general compact group
the construction reads the same: couple each state to a learned function of the attended
states and their successors, expanded over irreducible representations. Moving the
architecture to another state manifold requires only the group's irreducible
representations in place of the harmonics and the same delay term. For motion data the
manifold would be a rotation group. These instantiations are future work.
\section{Experiments}
\label{sec:experiments}

\subsection{Protocol}
\label{sec:protocol}

All comparisons use identical recipes: the same data order, optimizer, learning rate,
schedule, batch size, and sequence length, with parameter counts matched to a common
one-million target. Realized counts differ by under four percent in directions that vary
by corpus. Section~\ref{sec:architecture} reports them exactly, and
Appendix~\ref{app:config} lists the complete resolved configuration of every main run.
The transformer baseline is a pre-norm RoPE transformer with SwiGLU feed-forward blocks,
the lowest-validation configuration of its family at this scale in our tuning. Model
selection uses validation loss only. Table~\ref{tab:main} compares models at a matched
budget of thirty epochs, which measures sample efficiency at matched parameters, and also
reports the transformer trained to convergence at fifty epochs (best validation over five
seeds) as a reference bar, alongside the FSN's converged fifty-epoch values on enwik8 in
the same block.

\subsection{Language and code modeling at matched parameters}

\begin{table}[t]
\centering
\caption{Validation bits per character at the one-million-parameter scale after thirty
epochs of identical training. The enwik8 column reports mean $\pm$ sample standard
deviation over five seeds for the FSN row and over three seeds for the transformer,
no-phase, and winding-register rows; the FSN-MF row's third seed diverged at epoch 17
and it therefore reports two seeds. The codeparrot column (one hundred million
characters of Python source, sequence length 256) reports the transformer as mean $\pm$
standard deviation over three seeds and the FSN rows at seed 0; the FSN's remaining code
seeds are discussed in the text. The bottom block reports
fifty-epoch converged values on enwik8: the transformer reference bar (best validation
over five seeds) and the FSN's three completed fifty-epoch seeds.}
\label{tab:main}
\begin{tabular}{lcc}
\toprule
Model & enwik8 & codeparrot \\
\midrule
Transformer (RoPE, SwiGLU) & \xfThirty\,$\pm$\,\xfThirtyStd & \xfCodeThirty\,$\pm$\,\xfCodeThirtyStd \\
FSN & \sfnThirty\,$\pm$\,\sfnThirtyStd & \sfnCodeThirty \\
FSN, no phases & \nophaseThirty\,$\pm$\,\nophaseThirtyStd & \nophaseCodeVal \\
FSN-MF (no MLP anywhere; $n{=}2$) & \mfThirty\,$\pm$\,\mfThirtyStd & --- \\
FSN-MF, winding register & \mfwThirty\,$\pm$\,\mfwThirtyStd & --- \\
\midrule
Transformer, converged (50 ep) & \xfConverged & --- \\
FSN, converged (50 ep) & \sfnConverged\,$\pm$\,\sfnConvergedStd & --- \\
\bottomrule
\end{tabular}
\end{table}

Table~\ref{tab:main} contains the main comparison. The FSN and its no-phase configuration
are reported as co-equal main models, because at the thirty-epoch budget the two are
statistically indistinguishable (Section~\ref{sec:ablations}). On enwik8 the epoch-matched
ordering is consistent across seeds and epochs. The validation loss of both main models is
below the transformer baseline's at every epoch from the first onward on every seed. On
code the same every-epoch ordering holds for two of the three FSN seeds, while the third
trained to a worse plateau (\sfnCodeSOne, against \sfnCodeThirty{} and \sfnCodeSTwo{} for
the other two seeds). The thirty-epoch models are also ahead of the converged baseline.
Every FSN seed and every no-phase seed is, by epoch 30, already below the \emph{fifty-epoch
converged} transformer reference of \xfConverged{} (the FSN seeds cross it at epochs
\sfnCrossEpochs), and the FSN's completed fifty-epoch runs converge to
$\sfnConverged \pm \sfnConvergedStd$, so at this scale the converged FSN also beats the
converged baseline at an equal budget. The no-phase configuration reaches the converged
fifty-epoch validation loss of the base Kuramoto attention model ($\tanConverged$) within
ten epochs on its fastest seed and within eleven on all three.

The two mean-field rows quantify the cost of removing every multilayer perceptron from
the stack. The FSN-MF finishes \mfGapThirty{} bits per character behind the epoch-matched
transformer and the winding-register variant (Appendix~\ref{app:config}) \mfwGapThirty{} behind, after leading it through the
first five and seven epochs respectively at the matched seed, and one of the three plain
FSN-MF seeds diverged at epoch 17. At this scale this family tracks the transformer without
beating it. The full FSN's margin
over the two mean-field rows, $\sfnMfGap$ and $\sfnMfwGap$ bits per character, measures how
much the feed-forward block contributes beyond the oscillator stack (Section~\ref{sec:ffn}).

Source code is a second domain because it repeats identifiers and structural patterns at
far higher density than encyclopedic text. The code result is positive but no larger than
the enwik8 result. On the two well-trained seeds the FSN is ahead of the transformer at
every epoch, and its epoch-30 value (\sfnCodeThirty, with a best of \sfnCodeBest{} at
epoch 29) is below the best value the transformer reaches at any point in its own thirty
epochs on any seed, but the margin is comparable to the enwik8 margin. Code at sequence
length 256 may be context-starved, so we re-ran the seed-0 pair at sequence length 512.
Doubling the context improves both models substantially, by \codeContextDropFSN{} bits per
character for the FSN and \codeContextDropXF{} for the transformer at epoch 30, far more
than the architecture choice contributes on this corpus. The final-epoch margin is in fact
smaller at 512 than at 256, $\codeMarginFiveTwelve$ versus $\codeMarginTwoFiftySix$
(epoch-mean $\codeMarginFiveTwelveMean$ and $\codeMarginTwoFiftySixMean$ over the
thirty epochs). The code result is therefore supported by the copy-depth evidence of
Section~\ref{sec:capability} rather than by context scaling. In the completed 512
comparison the FSN finishes at \sfnCodeFiveTwelve{} against the transformer's thirty-epoch
\xfCodeFiveTwelve, a margin of $\codeMarginFiveTwelve$ bits per character. Repeating the
per-depth decomposition of Section~\ref{sec:capability} on the code validation split,
against the three-seed mean of the transformer's per-token cross-entropy, the advantage of
the two healthy seeds grows monotonically with copy depth across the four bins of depth
four and beyond, reaching between $\cdcDeepLo$ and $\cdcDeepHi$ bits per character on the
two deepest bins with window-cluster confidence intervals clear of zero. Such deep
positions are far more common on code: \codeDeepShare{} percent of validation tokens sit at
copy depth twenty-four or beyond, against \enwikDeepShare{} percent on enwik8. Even the
plateau-failure seed, though $\cdcFailZeroOne$ bits per character behind the transformer on
the shallowest bin, retains advantages of $\cdcFailSixteen$ and $\cdcFailTwentyFour$ on the
two deepest bins with confidence intervals clear of zero, so its deficit is general rather
than specific to copying. Appendix~\ref{app:copydepth} reports the full code table
alongside the enwik8 results.

The training failures differ in how early they become detectable. Both code plateau
failures separated from their sibling seeds at the first epoch, one-thirtieth of the
training budget. The failed sequence-length-256 seed read \codeFailEpOne{} bits per
character at epoch one against \codeHealthyEpOneA{} and \codeHealthyEpOneB{} for its
siblings, and the plateauing seed of a sequence-length-512 replication arm (whose other
seeds remain incomplete) read \codeFiveTwelveFailEpOne{} against
\codeFiveTwelveHealthyEpOneA{} and \codeFiveTwelveHealthyEpOneB{}, separations of more than
\codeSeedSepMin{} bits per character at roughly thirty times the spread between the healthy
siblings in each case. The FSN-MF divergence gave no early warning: that seed was
indistinguishable from its siblings through epoch sixteen, where its value of
\mfDivergeEpSixteen{} sat between the other two, and it produced a non-finite loss suddenly
at epoch seventeen. Three failures are too few to validate an early-restart policy.

\subsection{Capability: long-range copying on natural text}
\label{sec:capability}

We test whether the frustrated coupling law improves retrieval-to-continuation by
decomposing enwik8 validation cross-entropy by copy depth. A
position is assigned depth $\ell$ when the longest match between the suffix ending at that
position and any earlier substring of its context window has length $\ell$, so positions of
high depth are the positions most directly served by in-context copying. Depth measures the
length of the match, not the distance back to the matching occurrence, which this analysis
does not condition on. Positions are pooled into six depth bins (Appendix~\ref{app:copydepth}
gives the full methodology and the complete table). The fixed reference for every
comparison is the converged fifty-epoch transformer baseline at its best epoch, and each
evaluated model's margin on a bin is the paired per-token cross-entropy difference in bits
between that model and the reference on identical tokens, with window-cluster bootstrap
ninety-five-percent confidence intervals. Negative margins favor the evaluated model. Every
comparison is replicated over three seeds. We evaluate the three converged seeds of the
base Kuramoto attention layer (best epochs \cdBaseEpochs) and the three converged seeds of
the FSN (best epochs \cdSfnEpochs) against this reference, reporting seed means with the
spread across seeds.

All three base-layer seeds are behind the converged transformer in every depth bin,
including the shallow bins where no copying mechanism is involved, and the deficit of every
seed peaks on the long-copy bins. On the sixteen-to-twenty-three bin the three seeds lose
between $\cdBaseSixteenMin$ and $\cdBaseSixteenMax$ bits per character to the transformer,
with a seed mean of $\cdBaseSixteenMean$, more than five times the seed-mean deficit on any
other bin, while on the deepest bin (depth twenty-four to thirty-two) the seed-mean deficit
returns to $\cdBaseTwentyFourMean$. The per-depth curve in Appendix~\ref{app:copydepth}
shows that this sixteen-to-twenty-three deficit is not a uniform law of depth but
concentrates in spikes at specific depths, the largest being $\cdCurveSeventeen$ bits per
character at depth seventeen, so the bin records a failure on specific long
repeated-content windows in the validation data. The base layer fails most severely on the
positions whose context already contains a long verbatim continuation, evidence for missing
copy behavior on natural text.

\begin{figure}[t]
\centering
% Source: figures/fig_copydepth.pdf built by figures/make_fig_copydepth.py (seed-mean
% rebuild 2026-06-12) from results/exploratory_figs/{copydepth_battery_ce_cache.npz,
% copydepth_seeds_ce_cache.npz, h_copy_enrich_depth_cache.npz}; three seeds per family
% (base comp_tan_1m_s0/s1/s2 ep46/49/49, FSN sakanoise50_s0/s1/s2 ep48/50/50) vs
% comp_xf_1m_s0 ep48; per-seed values cross-checked against copydepth_stats.out.
\includegraphics[width=\linewidth]{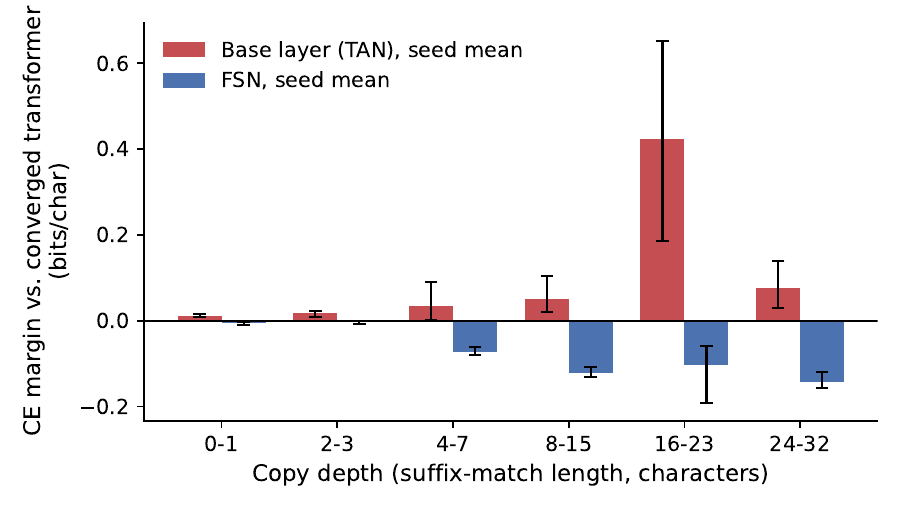}
\caption{Per-token cross-entropy margins against the converged transformer baseline, in
bits per character, by copy-depth bin, for the converged base Kuramoto attention layer
and the converged FSN. Each bar is the mean over the three seeds of that model, and the
whiskers span the per-seed range, that is, they run from the minimum to the maximum of
the three per-seed margins in the bin. Bars below the zero line indicate an advantage
over the transformer. Every base-layer seed is behind the transformer in every bin, with
the failure concentrated on the sixteen-to-twenty-three bin. Every FSN seed is ahead of
the transformer on every bin of depth four and beyond. Per-seed values with confidence
intervals are in Table~\ref{tab:copydepth}.}
\label{fig:copydepth}
\end{figure}

The FSN closes the deficit on the same bins against the same reference. All three FSN seeds
are ahead of the converged transformer on every bin of depth four and beyond, and on every
one of those bins the confidence interval excludes zero for every seed. The seed-mean
advantages are $\cdSfnFourSevenMean$ bits per character on the four-to-seven bin and
$\cdSfnEightFifteenMean$ on the eight-to-fifteen bin, where the per-seed spread is tight
(ranges of $\cdSfnFourSevenRange$ and $\cdSfnEightFifteenRange$ respectively), and
$\cdSfnTwentyFourMean$ on the twenty-four-to-thirty-two bin (range $\cdSfnTwentyFourRange$).
The sixteen-to-twenty-three bin has a seed-mean advantage of $\cdSfnSixteenMean$ with a
wider spread (range $\cdSfnSixteenRange$), because one seed is markedly further ahead of
the transformer there than the other two, though every seed is ahead. The advantage is
concentrated on the long-range copy events associated with the anticipation mechanism of
Section~\ref{sec:theory}. On the shallow bins the FSN's advantage is marginal, with seed
means of $\cdSfnTwoThreeMean$ on the two-to-three bin and $\cdSfnZeroOneMean$ on the
zero-to-one bin and with per-seed confidence intervals that touch or cross zero on several
seeds, so the FSN's win over the converged baseline is localized at deep-copy bins rather
than uniform. The no-phase configuration, evaluated at its own best epoch on each of its
three seeds, tracks the FSN closely on every bin of depth four and beyond
(Appendix~\ref{app:copydepth}), so removing the static phases leaves the deep-copy effect
intact. Figure~\ref{fig:copydepth} plots the base-layer and FSN seed means.

This decomposition has a precedent at larger scale. \citet{arora2024zoology} classify
subword tokens whose preceding bigram has already appeared in the context as
associative-recall hits and report that this slice accounts for most of the
validation-loss gap between attention models and gated-convolution models on the Pile.
The positions that the depth bins isolate are therefore also the positions on which
attention has been observed to earn its advantage over architectures that replace it,
and the FSN's advantage over the converged transformer is concentrated on the same part
of the distribution.

The deep-copy advantage also appears before convergence: an FSN-family checkpoint trained for only eleven
epochs is already ahead of the same converged reference on every bin of depth four and
beyond, with margins of $\copyFourSeven$, $\copyEightFifteen$, $\copySixteen$, and
$\copyTwentyFour$ bits per character on the four deep bins in depth order.

\subsection{Mechanism: reading the trained kernel}
\label{sec:mechanism}

\begin{figure}[t]
\centering
% Source: results/daido_probe/sakanoise50_s0/best.pt (best epoch 48);
% built by figures/make_fig_coupling.py from fused_w0/fused_w1 [4 layers, 176 coords, 3 harmonics].
\includegraphics[width=\linewidth]{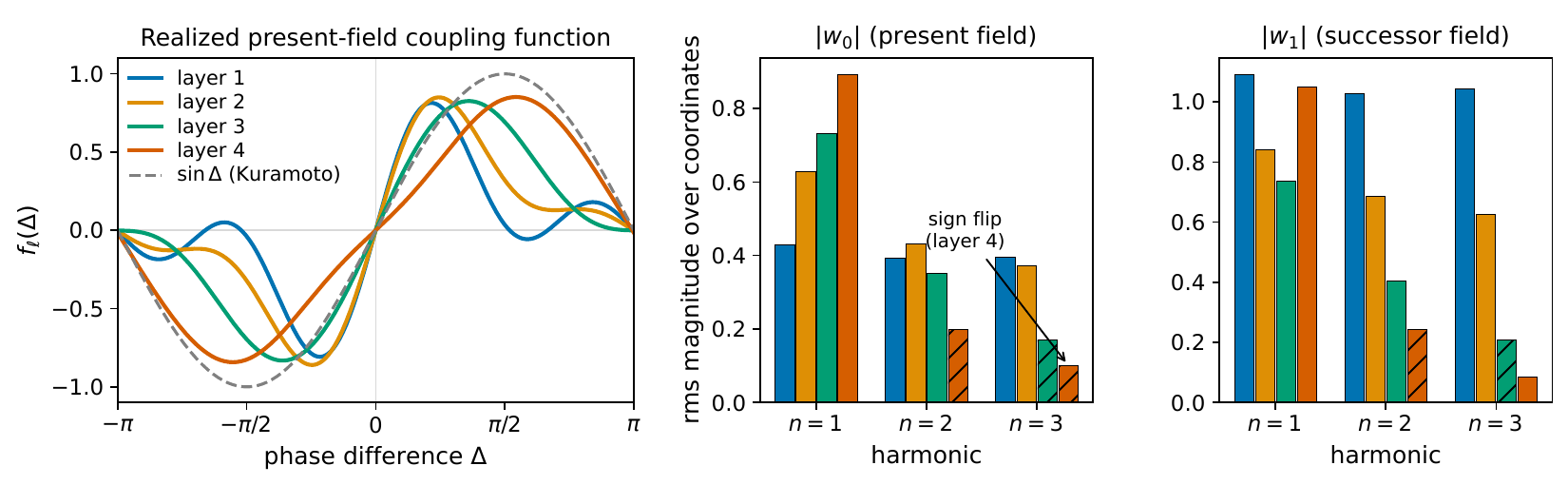}
\caption{The trained kernel, read as coupling functions. Left: the realized present-field
coupling function of each layer,
$f_\ell(\Delta)=\frac{1}{k}\sum_{a}\sum_{n=1}^{3}\bigl[\mathrm{Re}\,w_0(\ell,a,n)\sin n\Delta
+\mathrm{Im}\,w_0(\ell,a,n)\cos n\Delta\bigr]$, with the Kuramoto reference $\sin\Delta$
dashed in gray. Right: root-mean-square over coordinates of the complex harmonic magnitudes
of the present field ($w_0$) and the successor field ($w_1$), by harmonic and layer.
Hatched bars mark harmonics whose coordinate-mean real coefficient trained to a negative
value; the annotation marks the layer-four third-harmonic sign flip. The first harmonic
strengthens with depth on the present field and the higher harmonics of both fields are
pruned with depth, concentrating multi-harmonic structure in the early layers.}
\label{fig:coupling}
\end{figure}

Every kernel parameter is a coefficient of a coupling function, so the trained
coefficients reconstruct the realized per-layer coupling functions in
Figure~\ref{fig:coupling}. Every layer trains an attractive first harmonic, and the first
harmonic of the present field strengthens monotonically with depth while the second and
third harmonics weaken, so the deepest layer realizes the coupling function closest to a
pure Kuramoto sine. The early layers instead spread comparable weight across all three
harmonics, producing the strongly non-sinusoidal profiles in the left panel. The sign
structure that Section~\ref{sec:theory} associates with repulsive transport appears in the
last two layers at small magnitude: the coordinate-mean real coefficient of the third
harmonic flips negative at layers three and four, and that of the second harmonic flips
negative at layer four, while the imaginary parts stay near zero throughout, so the
realized phase offsets sit at zero or $\pi$ rather than at intermediate values. The
successor field keeps a strong first harmonic at every layer while the model prunes the
delay term's higher harmonics with depth, placing multi-harmonic shaping in the present
field of the early layers and anticipation in the first harmonic of the successor field.
The signed value gates train to a stable mixture of attractive and repulsive coordinates.

\subsection{A scale ladder}
\label{sec:ladder}

A parameter-doubling ladder (one, two, four, and eight million parameters; same recipe,
seed 0) tests whether the epoch-matched enwik8 advantage persists with scale. At two
million parameters the FSN finishes at \ladderTwoFSN{} against the transformer's completed
thirty-epoch value of \ladderTwoXFFinal. At four million the FSN finishes at \ladderFourFSN{}
against the transformer's \ladderFourXFFinal. At eight million the FSN reaches
\ladderEightFSN{} by epoch \ladderEightFSNEpoch{}, below the transformer's
epoch-\ladderEightFSNEpoch{} value (\ladderEightXF{}) and below its completed thirty-epoch
value (\ladderEightXFFinal). Parameter matching on the ladder is imperfect in the
conservative direction. The realized FSN models are lighter than their transformer partners
by $\ladderTwoLighter$, $\ladderFourLighter$, and $\ladderEightLighter$ percent at the
two-, four-, and eight-million rungs respectively, so the FSN side of each pair is slightly
handicapped. The epoch-matched advantage does not shrink with scale across the ladder.

\subsection{Target-loss wall-clock}
\label{sec:wallclock}

\begin{figure}[t]
\centering
\includegraphics[width=\linewidth]{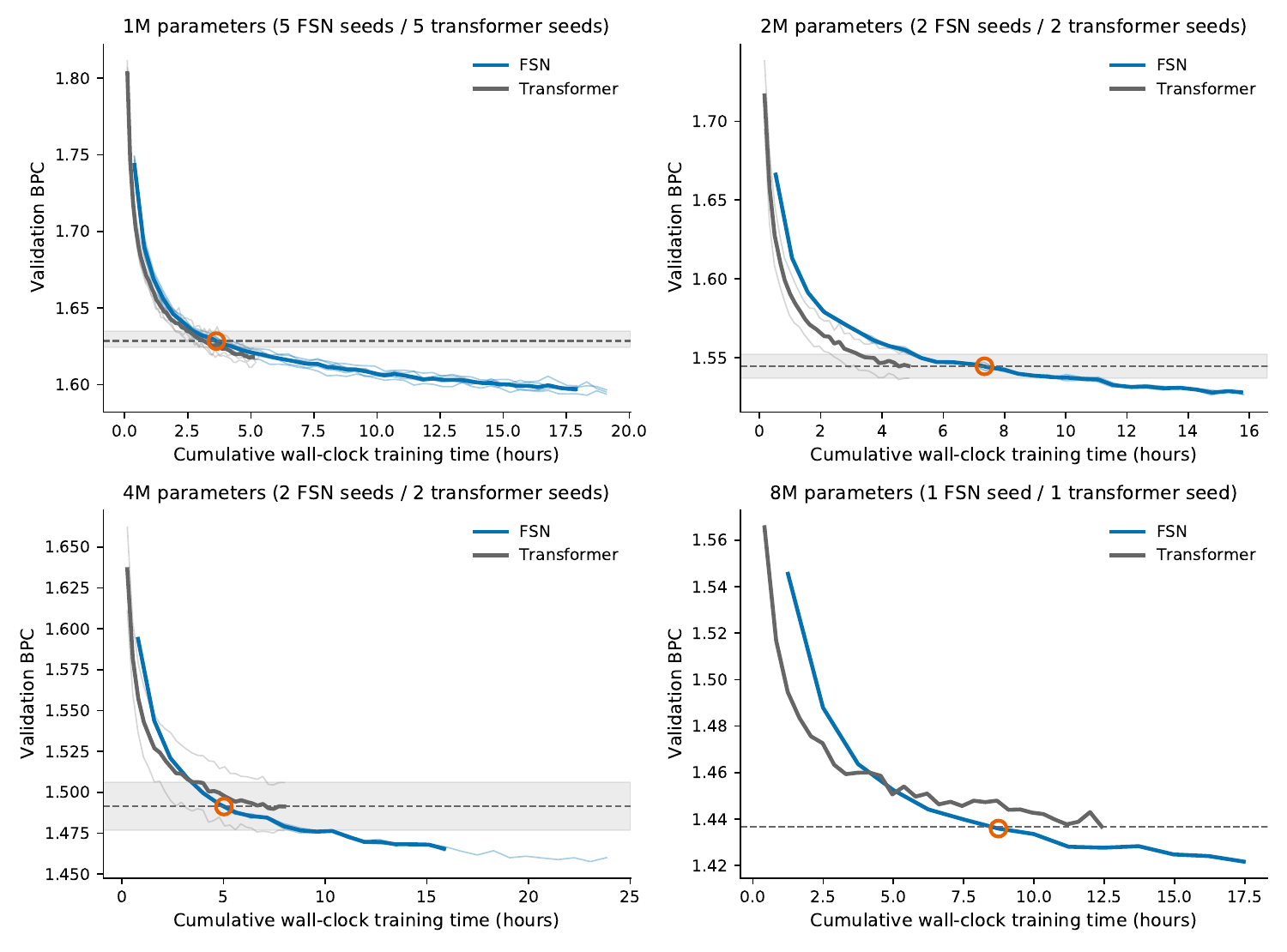}
\caption{Validation loss against cumulative wall-clock training time at each rung of the
parameter ladder on enwik8. Faint lines are individual seeds (FSN in blue, transformer in
gray); bold lines are the seed means. The shaded band spans the transformer seeds'
thirty-epoch validation losses and the dashed line their mean; the open circle marks where
the mean FSN curve crosses that mean target. Seed counts are given in each panel title.}
\label{fig:wallclock}
\end{figure}

The current implementation trains roughly three times more slowly per epoch than the
transformer at every scale ($\wcSlowOne\times$, $\wcSlowTwo\times$, $\wcSlowFour\times$, and
$\wcSlowEight\times$ at one, two, four, and eight million parameters).
Table~\ref{tab:wallclock} sets this against the FSN's faster convergence. For each transformer
seed we record its thirty-epoch validation loss and the cumulative train-plus-validation
wall-clock to reach it, and for each FSN seed the wall-clock at which it first reaches that
loss. Their ratio is the wall-clock cost of matching that transformer's quality. Against the
mean transformer the FSN reaches the thirty-epoch loss in $\wcMeanOne\times$ and
$\wcMeanTwo\times$ the transformer's wall-clock at one and two million parameters and in
$\wcMeanFour\times$ at four and eight million. Against the lowest-loss transformer seed at each
scale, the hardest target, the ratios are $\wcBestOne\times$, $\wcBestTwo\times$,
$\wcBestFour\times$, and $\wcBestEight\times$. Figure~\ref{fig:wallclock} plots validation
loss against cumulative wall-clock at each scale, with the FSN crossing of the mean transformer
loss marked. On code the per-epoch picture is similar: the seed-level ratios sit at parity at
sequence length 256 (\wallCodeRatioA{} and \wallCodeRatioB{} on the two well-trained seeds) and
move against the FSN at sequence length 512 (\wallCodeFiveTwelveRatio, at a
\wallCodeFiveTwelveSlowdown$\times$ per-epoch overhead).

At the crossing the two trajectories are at very different stages. When the FSN first
reaches the transformer's final validation loss the transformer has essentially stopped
improving, gaining at most about one millibit per character per epoch over its final five
epochs at every rung of the ladder. The FSN passes through the same loss value while still
improving at rates between roughly one and nine millibits per character per epoch, and on
the completed thirty-epoch rungs continues after the crossing to a final loss between
\crossAfterMin{} and \crossAfterMax{} bits per character below the target. No transformer
run reaches the final loss of any FSN run at any scale or seed we trained.

\begin{table}[t]
\centering
\caption{Cross-seed wall-clock at the thirty-epoch target on enwik8. For each transformer
seed we take its thirty-epoch validation loss and its cumulative train-plus-validation
wall-clock to that epoch; for each FSN seed we take the wall-clock at which it first reaches
that loss. Each ratio is the FSN wall-clock divided by the transformer wall-clock, reported
as the median over FSN seeds with the range across seeds. ``Mean transformer'' uses the mean
thirty-epoch loss and mean wall-clock; ``lowest-loss transformer'' uses the transformer seed
with the lowest thirty-epoch loss. Per-epoch slowdown is the ratio of mean FSN to mean
transformer wall-clock per epoch.}
\label{tab:wallclock}
\small
\setlength{\tabcolsep}{5pt}
\begin{tabular}{lcccc}
\toprule
Params & seeds (FSN\,$\times$\,XF) & per-epoch slowdown & vs.\ mean transf. & vs.\ lowest-loss transf. \\
\midrule
1M & $5\times5$ & $\wcSlowOne\times$ & $\wcMeanOne\times$ [\wcMeanOneLo, \wcMeanOneHi] & $\wcBestOne\times$ [\wcBestOneLo, \wcBestOneHi] \\
2M & $2\times2$ & $\wcSlowTwo\times$ & $\wcMeanTwo\times$ [\wcMeanTwoLo, \wcMeanTwoHi] & $\wcBestTwo\times$ [\wcBestTwoLo, \wcBestTwoHi] \\
4M & $2\times2$ & $\wcSlowFour\times$ & $\wcMeanFour\times$ [\wcMeanFourLo, \wcMeanFourHi] & $\wcBestFour\times$ [\wcBestFourLo, \wcBestFourHi] \\
8M & $1\times1$ & $\wcSlowEight\times$ & $\wcMeanEight\times$ & $\wcBestEight\times$ \\
\bottomrule
\end{tabular}
\end{table}

\subsection{Limitations of the present evidence}
\label{sec:limitations-exp}

The main comparisons are at the one-million-parameter scale on character-level corpora at a
thirty-epoch budget. Converged fifty-epoch values are reported on enwik8 over three seeds,
and the scale ladder runs to eight million parameters. Converged comparisons for the
no-phase, mean-field, code, and ladder arms, and subword-tokenized corpora, are left to
future work. One FSN code seed and one FSN-MF seed trained poorly or diverged, so seed robustness
outside enwik8 is not yet established. The current
% verifier 2026-06-12: measured per-epoch slowdowns are 3.6-3.7x at 1M (Appendix B),
% 3.2x at 2M, 3.0x at 4M/8M, 3.9x at code seq 512; "two to three" understated the cost.
implementation trains at roughly three to four times the wall-clock cost of the baseline
per epoch. Against the mean transformer the FSN reaches the thirty-epoch loss in less
wall-clock at four and eight million parameters and in more at one and two million
(Table~\ref{tab:wallclock}). Appendix~\ref{app:efficiency} discusses the measured
implementation costs.
\section{Mechanism Ablations}
\label{sec:ablations}

The converged comparisons of Section~\ref{sec:experiments} already isolate two components of the
coupling law. Removing the static frustration phases gives the no-phase configuration, which at
thirty epochs is statistically indistinguishable from the full FSN ($\nophaseThirty \pm
\nophaseThirtyStd$ over three seeds against $\sfnThirty \pm \sfnThirtyStd$ over five). Early in
training the phases do help, with epoch-2 seed means of $\sfnEpTwoMean$ against $\nophaseEpTwoMean$
over the three matched seeds, but the lead closes by convergence, so the static phases buy
early-training speed rather than converged quality.

Removing the feed-forward block gives the FSN-MF variant, which costs $\sfnMfGap$ bits per character
at thirty epochs (Table~\ref{tab:main}). Section~\ref{sec:ffn} attributes this gap to the winding of
the phase state, which the SwiGLU reads and a phasor block cannot.

The remaining terms we can check only at screen level. In a two-epoch single-seed screen, removing
the delay term and leaving the harmonics alone costs $\ablNoSuccDelta$ bits per character, the
largest single subtraction we measured and consistent with the role of anticipation in
Section~\ref{sec:theory}. A converged, multi-seed ablation of the delay term is left to future work.
The transport carried by the higher harmonics, the repulsion carried by the signed value gates, and
the per-coordinate structure of the kernel are visible in the trained coefficients of
Section~\ref{sec:mechanism}, but controlled subtractions of them remain to be run.

\section{Related Work}
\label{sec:related}

\paragraph{Associative-memory and dynamical readings of attention.} The modern Hopfield
update is equivalent to transformer attention, giving one account of attention as
content-addressed associative retrieval \citep{ramsauer2021hopfield}. Iterating attention
through depth admits a complementary, dynamical reading. With residual and feed-forward
paths removed, pure self-attention loses rank doubly exponentially and collapses toward
token uniformity~\citep{dong2021attention}. Interacting-particle analyses of simplified
self-attention dynamics show asymptotic clustering, with the limiting behavior controlled by
the value matrix~\citep{geshkovski2023emergence,geshkovski2023mathematical}, and related
work proves convergence toward consensus-like limits under various
assumptions~\citep{abella2025consensus}. Analyses of attention masks, LayerNorm, and causal
masking show that collapse persists in broad masked settings while normalization, value
matrices, sparsity, and causal structure reshape the
limits~\citep{wu2024attentionmasks,karagodin2024clustering}. These results motivate the
separation used here between a score map that retrieves relevant states and a value pathway
that determines the dynamical operation applied to them.

\paragraph{Oscillator readings of attention.} Kuramoto attention~\citep{nunley2026kuramoto}
supplies the pure-attraction control, a torus-valued attention layer whose score map is a
gated phase-coherence query-key mechanism and whose value update is adaptive Kuramoto
coupling. The FSN keeps that state space and score map, changes the coupling function, and
measures the resulting capability differences. Artificial Kuramoto oscillatory
neurons~\citep{miyato2024akorn} add Kuramoto dynamics to standard networks as an auxiliary
binding and robustness mechanism, whereas the FSN makes oscillator dynamics the attention
computation itself.

\paragraph{Synchronization theory.} The frustration vocabulary comes from the
Kuramoto--Sakaguchi model~\citep{sakaguchi1986soluble}, and the harmonic generalization of
the coupling function is due to Daido~\citep{daido1996onset}. Anticipating synchronization
through delay coupling, introduced by Voss~\citep{voss2000anticipating}, is the precedent for
the FSN's coupling of each token to the successor of an attended token. The FSN-MF feed-forward block couples
through collective modes in the mean-field tradition of the same literature, and its
multi-oscillator interactions are higher-order generalizations of pairwise
coupling~\citep{bick2023higher}.

\paragraph{Mechanistic accounts of in-context behavior.} The copy-depth measurements in
Section~\ref{sec:capability} target the behavior attributed to induction
heads~\citep{olsson2022context}, the circuit family responsible for in-context copying in
transformers. The token slice they isolate follows the associative-recall analysis of
\citet{arora2024zoology}, which classifies a token as an associative-recall hit when the
bigram it forms with its predecessor has already appeared in the context, and reports that
this slice, roughly six percent of tokens, accounts for most of the validation-loss gap
between attention models and gated-convolution models on the Pile. The copy-depth bins of
Section~\ref{sec:capability} generalize that single repeated bigram to graded copy depth.
The FSN is a non-transformer datapoint for this literature. An architecture can have full
attention yet still lag a matched transformer on long repeated contexts when its value
pathway is purely attractive, and restoring a transition-reading pathway removes the lag.

\paragraph{Small character-level sequence models.} enwik8 is a common testbed for
architecture changes in sequence models, including Transformer-XL and other long-context
variants~\citep{dai2019transformerxl,rae2020compressive}. The closest precedent is
Group-TransformerXL, which reports parameter-matched improvements over Transformer-XL at
roughly four and eight million parameters on enwik8 and text8~\citep{park2020group}.
Recurrent Memory Transformer improves memory efficiency and long-range synthetic tasks and
reports language-modeling quality comparable to Transformer-XL using fewer memory
states~\citep{bulatov2022rmt}. Like these models, the FSN improves on a transformer-family
baseline on enwik8, and the cross-seed wall-clock comparison is in
Section~\ref{sec:wallclock}.

\paragraph{Position and gating.} The score function inherits rotary position
embeddings~\citep{su2021roformer}, which become literal phase drifts on the torus, and the
full FSN retains the standard SwiGLU feed-forward block~\citep{shazeer2020glu}, which holds
the universal-cover channel of Section~\ref{sec:ffn}.
\section{Conclusion}
\label{sec:conclusion}

Associative-memory, rank-collapse, and interacting-particle views of attention separate
retrieval from the operation applied after retrieval. The Frustrated Synchronization
Network makes that operation a learned coupling law. Equation~\ref{eq:kernel} defines
attention as a learned coupling over phase states, with harmonic terms that transport states
between configurations and a one-step delay term that anticipates. The delay term is identical
to Kuramoto--Sakaguchi coupling frustrated by the data's own transitions
(Proposition~\ref{prop:sakaguchi}), so the same score map that retrieves analogous contexts
can move the current token toward what followed them. The attention pathway is directly
inspectable. Every coefficient of the coupling kernel is a named object of an existing
mathematical literature, and the trained models read, coefficient by coefficient, as coupling
functions.

We compare epoch-matched and converged models at one million parameters on character-level
text and code, scale to eight million parameters, and measure copy depth to distinguish
retrieval-as-consensus from retrieval-to-continuation. Subword tokenization and larger scales
are left to future work.

The kernel construction lifts to any compact group, with irreducible representations in place
of harmonics, so it extends to rotation groups for motion data and beyond. Because phases,
frustrations, delays, and mean-field coupling are native operations for physical oscillator
hardware, the FSN family can also be studied as a compilation target for such substrates. A
fully periodic shared-bank model, with all phases on one shared oscillator bank, is the
natural endpoint of that hardware direction and remains future work.

\bibliographystyle{plainnat}
\bibliography{references}

@article{nunley2026kuramoto,
  author  = {Nunley, Joshua},
  title   = {{Kuramoto} Attention: Self-Attention as Adaptive Coupling on the Torus},
  journal = {arXiv preprint},
  year    = {2026}
}

@article{sakaguchi1986soluble,
  author  = {Sakaguchi, Hidetsugu and Kuramoto, Yoshiki},
  title   = {A Soluble Active Rotator Model Showing Phase Transitions via Mutual Entrainment},
  journal = {Progress of Theoretical Physics},
  volume  = {76},
  number  = {3},
  pages   = {576--581},
  year    = {1986}
}

@article{daido1996onset,
  author  = {Daido, Hiroaki},
  title   = {Onset of Cooperative Entrainment in Limit-Cycle Oscillators with Uniform All-to-All Interactions: Bifurcation of the Order Function},
  journal = {Physica D},
  volume  = {91},
  number  = {1--2},
  pages   = {24--66},
  year    = {1996}
}

@article{voss2000anticipating,
  author  = {Voss, Henning U.},
  title   = {Anticipating Chaotic Synchronization},
  journal = {Physical Review E},
  volume  = {61},
  number  = {5},
  pages   = {5115--5119},
  year    = {2000}
}

@article{bick2023higher,
  author  = {Bick, Christian and Gross, Elizabeth and Harrington, Heather A. and Schaub, Michael T.},
  title   = {What Are Higher-Order Networks?},
  journal = {SIAM Review},
  volume  = {65},
  number  = {3},
  pages   = {686--731},
  year    = {2023}
}

@article{geshkovski2023mathematical,
  author  = {Geshkovski, Borjan and Letrouit, Cyril and Polyanskiy, Yury and Rigollet, Philippe},
  title   = {A Mathematical Perspective on Transformers},
  journal = {arXiv preprint arXiv:2312.10794},
  year    = {2023}
}

@inproceedings{dong2021attention,
  author    = {Dong, Yihe and Cordonnier, Jean-Baptiste and Loukas, Andreas},
  title     = {Attention is Not All You Need: Pure Attention Loses Rank Doubly Exponentially with Depth},
  booktitle = {Proceedings of the 38th International Conference on Machine Learning},
  pages     = {2793--2803},
  year      = {2021},
  volume    = {139},
  series    = {Proceedings of Machine Learning Research},
  publisher = {PMLR}
}

@inproceedings{geshkovski2023emergence,
  author    = {Geshkovski, Borjan and Letrouit, Cyril and Polyanskiy, Yury and Rigollet, Philippe},
  title     = {The Emergence of Clusters in Self-Attention Dynamics},
  booktitle = {Advances in Neural Information Processing Systems},
  year      = {2023}
}

@inproceedings{karagodin2024clustering,
  author    = {Karagodin, Nikita and Polyanskiy, Yury and Rigollet, Philippe},
  title     = {Clustering in Causal Attention Masking},
  booktitle = {Advances in Neural Information Processing Systems},
  volume    = {37},
  year      = {2024}
}

@inproceedings{ramsauer2021hopfield,
  author    = {Ramsauer, Hubert and Sch{\"a}fl, Bernhard and Lehner, Johannes and Seidl, Philipp and Widrich, Michael and Adler, Thomas and Gruber, Lukas and Holzleitner, Markus and Pavlovi{\'c}, Milena and Sandve, Geir Kjetil and Greiff, Victor and Kreil, David and Kopp, Michael and Klambauer, G{\"u}nter and Brandstetter, Johannes and Hochreiter, Sepp},
  title     = {Hopfield Networks is All You Need},
  booktitle = {International Conference on Learning Representations},
  year      = {2021}
}

@inproceedings{abella2025consensus,
  author    = {Abella, {\'A}lvaro Rodr{\'i}guez and Silvestre, Jo{\~a}o Pedro and Tabuada, Paulo},
  title     = {Consensus Is All You Get: The Role of Attention in Transformers},
  booktitle = {Proceedings of the 42nd International Conference on Machine Learning},
  pages     = {174--184},
  year      = {2025},
  volume    = {267},
  series    = {Proceedings of Machine Learning Research},
  publisher = {PMLR}
}

@inproceedings{wu2024attentionmasks,
  author    = {Wu, Xinyi and Ajorlou, Amir and Wang, Yifei and Jegelka, Stefanie and Jadbabaie, Ali},
  title     = {On the Role of Attention Masks and {LayerNorm} in Transformers},
  booktitle = {Advances in Neural Information Processing Systems},
  volume    = {37},
  year      = {2024}
}

@inproceedings{dai2019transformerxl,
  author    = {Dai, Zihang and Yang, Zhilin and Yang, Yiming and Carbonell, Jaime and Le, Quoc V. and Salakhutdinov, Ruslan},
  title     = {Transformer-{XL}: Attentive Language Models Beyond a Fixed-Length Context},
  booktitle = {Proceedings of the 57th Annual Meeting of the Association for Computational Linguistics},
  pages     = {2978--2988},
  year      = {2019}
}

@inproceedings{rae2020compressive,
  author    = {Rae, Jack W. and Potapenko, Anna and Jayakumar, Siddhant M. and Hillier, Chloe and Lillicrap, Timothy P.},
  title     = {Compressive Transformers for Long-Range Sequence Modelling},
  booktitle = {International Conference on Learning Representations},
  year      = {2020}
}

@inproceedings{park2020group,
  author    = {Park, Sungrae and Cha, Junbum and Kim, Geewook and Kim, Ji-Hoon and Lee, Junyeop and Lee, Hwalsuk},
  title     = {Scale Down Transformer by Grouping Features for a Lightweight Character-level Language Model},
  booktitle = {Proceedings of the 28th International Conference on Computational Linguistics},
  pages     = {6883--6893},
  year      = {2020}
}

@inproceedings{bulatov2022rmt,
  author    = {Bulatov, Aydar and Kuratov, Yuri and Burtsev, Mikhail S.},
  title     = {Recurrent Memory Transformer},
  booktitle = {Advances in Neural Information Processing Systems},
  year      = {2022}
}

@inproceedings{miyato2024akorn,
  author    = {Miyato, Takeru and Löwe, Sindy and Geiger, Andreas and Welling, Max},
  title     = {Artificial {Kuramoto} Oscillatory Neurons},
  booktitle = {International Conference on Learning Representations},
  year      = {2025}
}

@article{olsson2022context,
  author  = {Olsson, Catherine and Elhage, Nelson and Nanda, Neel and Joseph, Nicholas and others},
  title   = {In-Context Learning and Induction Heads},
  journal = {Transformer Circuits Thread},
  year    = {2022}
}

@article{su2021roformer,
  author  = {Su, Jianlin and Lu, Yu and Pan, Shengfeng and Murtadha, Ahmed and Wen, Bo and Liu, Yunfeng},
  title   = {{RoFormer}: Enhanced Transformer with Rotary Position Embedding},
  journal = {arXiv preprint arXiv:2104.09864},
  year    = {2021}
}

@article{shazeer2020glu,
  author  = {Shazeer, Noam},
  title   = {{GLU} Variants Improve Transformer},
  journal = {arXiv preprint arXiv:2002.05202},
  year    = {2020}
}

@inproceedings{arora2024zoology,
  author    = {Arora, Simran and Eyuboglu, Sabri and Timalsina, Aman and Johnson, Isys and Poli, Michael and Zou, James and Rudra, Atri and R{\'e}, Christopher},
  title     = {Zoology: Measuring and Improving Recall in Efficient Language Models},
  booktitle = {International Conference on Learning Representations},
  year      = {2024}
}

\appendix

\section{Configuration details}
\label{app:config}

Table~\ref{tab:config} lists the resolved configuration of the main runs, read from the
resolved configuration files written by the training pipeline at launch
(\texttt{config\_resolved.yaml} in each run directory) and from the run metrics. The FSN
columns are the runs behind Table~\ref{tab:main} (enwik8: \texttt{sakanoise50}; code:
\texttt{sakacode30}), and the transformer columns are their matched baselines
(\texttt{comp\_xf\_1m} and \texttt{xfmcode30}). The two architectures share every training
hyperparameter and differ only in the model block.

\begin{table}[t]
\centering
\caption{Resolved configuration of the main one-million-parameter runs. Rows in the
upper block describe the models; rows in the lower block are shared by both architectures
exactly. Parameter counts are the realized totals reported by the training pipeline.}
\label{tab:config}
\small
\begin{tabular}{lll}
\toprule
 & FSN & Transformer baseline \\
\midrule
Layers & 4 & 4 \\
State / width & $k=176$ phases per token & $d=120$ (enwik8), $d=124$ (code) \\
Attention & one causal score map per layer & 1 head \\
Position & RoPE as phase drift, score only & RoPE \\
Feed-forward & SwiGLU, hidden multiplier $2.0$ & SwiGLU, hidden multiplier $4$ \\
QK gates & shared pair, softplus, normalized & --- \\
Value gate & one shared linear (signed) gate & --- \\
Bounded update & norm-matched tanh, $\alpha$ init $2\pi$ & --- \\
Kernel & fused complex, $N=3$, per-coordinate & --- \\
Kernel init & noise init; successor gate init $1.5$ & --- \\
Readout & phase-coherence scores, learned temp. & linear head \\
Width selection & pinned at $k=176$ & matched to $10^6$ target \\
Parameters (enwik8) & \sfnParams & \xfParams \\
Parameters (code) & \sfnCodeParams & \xfCodeParams \\
\midrule
Optimizer & \multicolumn{2}{l}{AdamW, weight decay $0.01$, PyTorch default $(\beta_1,\beta_2)$} \\
Learning rate & \multicolumn{2}{l}{$10^{-3}$, constant (no warmup, no decay)} \\
Gradient clip & \multicolumn{2}{l}{global norm $1.0$} \\
Batch / sequence & \multicolumn{2}{l}{batch 64, sequence length 256 (512 in the long-context arm)} \\
Data windows & \multicolumn{2}{l}{training stride 64, evaluation stride 128} \\
Vocabulary & \multicolumn{2}{l}{character-level; 205 symbols (enwik8), 97 symbols (code)} \\
Dropout & \multicolumn{2}{l}{$0.1$} \\
Epochs & \multicolumn{2}{l}{30 (main budget), 50 (convergence tier)} \\
% verifier 2026-06-12: FSN enwik8 30-ep row is now five seeds (0-4); see numbers.tex \sfnThirty.
Seeds & \multicolumn{2}{l}{0--2 (FSN enwik8 30-ep row: 0--4); transformer converged bar over 0--4} \\
Hardware & \multicolumn{2}{l}{one A100-class GPU per run; fp32; no compilation} \\
\bottomrule
\end{tabular}
\end{table}

\paragraph{Gate parameterization.} The query and key gates of Equation~\ref{eq:score} are
a single pair of linear maps shared across all layers, applied to the trigonometric state
features $(\cos\theta, \sin\theta)$, with weights initialized to zero and biases to one, so
that the gates start uniform. Each gate passes through a softplus and is then normalized by
its mean across coordinates (with a floor of $10^{-6}$ on the denominator), so the gates
are nonnegative and average to one. The temperature $\tau$ is learned. The value gate is
likewise a single linear map shared across all layers, applied to the same state features
with the same zero-weight, unit-bias initialization, but its activation is the identity.
Leaving the value gate signed is what permits the per-coordinate attractive and repulsive
mixtures of Section~\ref{sec:mechanism}.

\paragraph{Bounded update.} Both the attention update and the feed-forward update pass
through the norm-matched tanh bound inherited from the base layer. The raw update keeps
its direction, and its per-token norm is rescaled to the norm of the reference
$\alpha \tanh(x)$, where $\alpha$ is a learned per-layer scalar initialized to $2\pi$.
The attention and feed-forward pathways use separate $\alpha$ parameters.
The bound is essential to stability. In a three-epoch screen with the bound removed
entirely, the run diverged to a validation loss of \bndNoneEpOne{} bits per character
within the first epoch.

\paragraph{Kernel.} The kernel of Equation~\ref{eq:kernel} is the fused complex
implementation with $N=3$ harmonics and per-coordinate coefficients, that is, independent
$w_0^{(n)}, w_1^{(n)} \in \C^{k}$ at every layer. The initialization is the noise variant
of the lean operating point. The first-harmonic real parts are set to
$w_1 = \sigma(1.5) \approx 0.82$ on the delayed field and $w_0 = 1 - \sigma(1.5) \approx
0.18$ on the present field, all other real parts are zero, and every imaginary part
receives independent Gaussian noise of standard deviation $0.05$. At exactly zero
imaginary part the loss surface is locally flat in the Sakaguchi phase directions, and the
small symmetry-breaking phases give those directions gradient from the first step.

\paragraph{Feed-forward variants.} Two further variants of the mean-field block of
Section~\ref{sec:ffn} map how much of the non-periodic lift the architecture must retain as it
moves toward fully periodic computation. The \emph{winding-register} variant gives the
mean-field block a small set of real gated modes that read the lifted phase state $\theta$
rather than the phasors $z$, with the mode budget rebalanced so that parameters still match,
and it has lower validation loss than the plain mean-field row (Table~\ref{tab:main}). The
\emph{dissolved} variant wraps the phases back into $(-\pi,\pi]$ after every layer and keeps
history in a small explicit register of accumulated updates, so that the state space is
$\mathbb{T}^k \times \mathbb{R}^{m}$ with $m \ll k$. Neither is part of the main FSN
comparison.

\paragraph{Matching and hardware.} The transformer's width is chosen by an automated
search (steps of four model dimensions) to land nearest the one-million-parameter target;
the FSN's width is pinned at $k=176$, chosen once against the same target and
reused unchanged across corpora and seeds. The realized totals, reported in
Section~\ref{sec:architecture}, differ because the two corpora have different vocabulary
sizes, which give the two architectures differently sized embedding tables. Every run in the
paper trains on a single A100-class GPU under SLURM, in full thirty-two-bit precision,
with no graph compilation, from the same training script. Seeds control model
initialization and data order jointly.

\section{Efficiency}
\label{app:efficiency}

All throughput figures are measured from run logs. Each run logs its training-loop token
throughput every epoch, and we report the mean of the per-epoch training values over all
completed epochs, at the shared batch size of 64 on one A100-class GPU per run.

At one million parameters on enwik8 the FSN trains at \tpsFSNOneM{} thousand tokens per
second against the transformer baseline's \tpsXFOneM{} thousand (three seeds each, with
per-seed means varying by under two percent for the FSN), a factor of \slowdownOneM. The code
corpus gives nearly identical figures, \tpsFSNCode{} against \tpsXFCode{} thousand. At
four million parameters the measured throughputs are \tpsFSNFourM{} against \tpsXFFourM{}
thousand tokens per second, a factor of \slowdownFourM. The wall-clock crossing analysis of
Section~\ref{sec:wallclock} is computed from these same per-epoch logs. The FSN-MF
trains at \tpsMFOneM{} thousand tokens per second at one million parameters. Peak
training memory at one million parameters is \memFSNOneM{}~GB for the FSN against
\memXFOneM{}~GB for the transformer. On the evaluation pass, where there is no backward
pass through the geometric update, the gap is smaller, \tpsEvalFSN{} million tokens per
second against \tpsEvalXF{} million.

The kernel is not the main source of the overhead. The no-phase configuration removes the
harmonic kernel entirely while keeping the rest of the stack, and it trains at
\tpsNophaseOneM{} thousand tokens per second, still a factor of \slowdownNophaseOneM{}
slower than the transformer. The full kernel therefore accounts for roughly
\kernelShare{} percent of the FSN's training step. The dominant overhead is the
bounded-update machinery inherited from the base layer, not the kernel this paper adds.
The delay axis of the kernel folds into the value tensor before the quadratic aggregation,
so the kernel's cost is additive rather than multiplicative. The delayed phasors are formed
by a one-position shift along the sequence and combined with the present field before the
attention product, adding work linear in sequence length, width, and harmonic count, while
the quadratic attention cost is paid exactly once.

These figures do not reflect an optimized implementation. Both architectures train in full
thirty-two-bit precision, without mixed precision and without compilation, from the same
generic training loop. Natural implementation work includes bfloat16 mixed precision and
graph compilation applied to both architectures under the same equal-treatment protocol as
every other comparison in the paper. We do not include projected speedups.

\section{Copy-depth methodology and full results}
\label{app:copydepth}

\paragraph{Depth assignment.} The copy-depth measurement of Section~\ref{sec:capability}
decomposes enwik8 validation loss by copy depth. A position is assigned depth $\ell$ when
the longest match between the suffix ending at that position and any earlier substring of
its context window has length $\ell$, capped at 32. Depth assignments are computed once
from the data alone, before any model is consulted, and positions are pooled into the
bins $0$--$1$, $2$--$3$, $4$--$7$, $8$--$15$, $16$--$23$, and $24$--$32$.

The depth assignment generalizes the associative-recall slice of
\citet{arora2024zoology}. Their analysis classifies a token as an associative-recall
hit when the bigram formed with its preceding token has already appeared in the
context, which corresponds to depth one in our assignment, and it excludes bigrams
that were common in the training data so that the slice is not dominated by memorized
pairs. Our assignment grades depth up to thirty-two rather than reducing it to a binary
condition at depth one, and uses characters rather than subword tokens. It applies no
training-frequency exclusion, because long character-level suffix matches are not common
training patterns, so the grading itself limits the contribution of memorized material at
every depth beyond the shallowest bins. The distance between a position and its matching
earlier occurrence is a second dimension of the same construction. Neither our analysis
nor the associative-recall analysis conditions on it, and a decomposition over depth and
distance together is left to future work.

\paragraph{Evaluation protocol.} The standard evaluation slice is the first 1{,}024
validation windows. Because deep-copy positions are rare there, a second slice of 192
windows, selected in advance for high deep-copy density, supplements the bins of depth
sixteen and beyond. The shallower bins use the standard slice alone. Every model is
evaluated on identical tokens. The margin of a model on a bin is the mean over the bin's
tokens of the paired per-token cross-entropy difference, in bits, between that model and
the fixed reference, which is the converged fifty-epoch transformer baseline at its best
validation epoch (epoch \cdXfEpoch). Negative margins favor the evaluated model.
Confidence intervals are window-cluster bootstrap ninety-five-percent intervals with
4{,}000 resamples, clustering tokens by their evaluation window. The six bins contain
\cdNtokZeroOne, \cdNtokTwoThree, \cdNtokFourSeven, \cdNtokEightFifteen, \cdNtokSixteen,
and \cdNtokTwentyFour{} tokens respectively. Every model family is evaluated at three
seeds, each seed at the best-validation checkpoint of its own run. For the converged
base Kuramoto attention layer the best epochs of seeds zero through two are
\cdBaseEpochs. For the converged FSN they are \cdSfnEpochs, and the three no-phase seeds
all reach their best validation at epoch \cdNophaseEpoch. Table~\ref{tab:copydepth}
reports the complete per-seed results, and Figure~\ref{fig:copydepth} plots the
base-layer and FSN seed means with whiskers spanning the per-seed range.

% CI cell helper for tab:copydepth: tiny window-cluster 95% interval under each margin.
\newcommand{\cic}[2]{{\tiny $[#1,#2]$}}

\begin{table}[t]
\centering
\caption{Per-seed copy-depth margins against the converged transformer baseline (best
epoch \cdXfEpoch), for all nine evaluated checkpoints, three seeds of each of the three
model families. Each cell in a seed row is the paired per-token cross-entropy difference
in bits per character, model minus transformer, on the bin's tokens, with the
window-cluster bootstrap ninety-five-percent confidence interval beneath it. The mean
rows give the seed mean of each family, with the range over the three seeds (maximum
minus minimum) beneath it. Negative values favor the evaluated model. Bins of depth
sixteen and beyond pool the standard and the deep-copy-enriched slices. Best epochs are
\cdBaseEpochs{} for base-layer seeds zero through two, \cdSfnEpochs{} for the FSN seeds,
and epoch \cdNophaseEpoch{} for all three no-phase seeds.}
\label{tab:copydepth}
\scriptsize
\setlength{\tabcolsep}{2pt}
% Source: per-seed margins and window-cluster 95% CIs (CI_win, 4000 resamples) from
% experiments/scripts/exploratory/copydepth_stats.py, full log
% experiments/results/exploratory_figs/copydepth_stats.out (2026-06-12); CE caches
% copydepth_battery_ce_cache.npz (seed 0) + copydepth_seeds_ce_cache.npz (seeds 1/2),
% reference comp_xf_1m_s0 ep48 asserted bit-identical across caches. Seed-0 margins are
% the \cd* macros of numbers.tex; seed-1/2 margins are in the numbers.tex seed-replication
% comment block; seed means and ranges are the \cd*Mean/\cd*Range macros. The seed-0 CIs
% supersede the earlier job-7369997 brackets (same centers; bootstrap RNG differs, so the
% interval endpoints move by less than 0.004).
\begin{tabular}{lrrrrrr}
\toprule
 & 0--1 & 2--3 & 4--7 & 8--15 & 16--23 & 24--32 \\
Tokens & \cdNtokZeroOne & \cdNtokTwoThree & \cdNtokFourSeven & \cdNtokEightFifteen & \cdNtokSixteen & \cdNtokTwentyFour \\
\midrule
Base layer, s0 & $\cdBaseZeroOne$ & $\cdBaseTwoThree$ & $\cdBaseFourSeven$ & $\cdBaseEightFifteen$ & $\cdBaseSixteen$ & $\cdBaseTwentyFour$ \\[-1pt]
 & \cic{+0.0054}{+0.0131} & \cic{+0.0129}{+0.0259} & \cic{+0.0041}{+0.0241} & \cic{+0.0043}{+0.0504} & \cic{+0.3929}{+0.4656} & \cic{+0.0163}{+0.0453} \\
Base layer, s1 & $+0.0108$ & $+0.0222$ & $+0.0895$ & $+0.1051$ & $+0.6525$ & $+0.1385$ \\[-1pt]
 & \cic{+0.0070}{+0.0147} & \cic{+0.0155}{+0.0291} & \cic{+0.0776}{+0.1024} & \cic{+0.0785}{+0.1320} & \cic{+0.6183}{+0.6839} & \cic{+0.1180}{+0.1611} \\
Base layer, s2 & $+0.0156$ & $+0.0095$ & $+0.0013$ & $+0.0202$ & $+0.1861$ & $+0.0628$ \\[-1pt]
 & \cic{+0.0118}{+0.0193} & \cic{+0.0030}{+0.0158} & \cic{-0.0076}{+0.0100} & \cic{-0.0006}{+0.0412} & \cic{+0.1647}{+0.2067} & \cic{+0.0364}{+0.0936} \\
Base layer, mean & $\cdBaseZeroOneMean$ & $\cdBaseTwoThreeMean$ & $\cdBaseFourSevenMean$ & $\cdBaseEightFifteenMean$ & $\cdBaseSixteenMean$ & $\cdBaseTwentyFourMean$ \\[-1pt]
 & {\tiny range $\cdBaseZeroOneRange$} & {\tiny range $\cdBaseTwoThreeRange$} & {\tiny range $\cdBaseFourSevenRange$} & {\tiny range $\cdBaseEightFifteenRange$} & {\tiny range $\cdBaseSixteenRange$} & {\tiny range $\cdBaseTwentyFourRange$} \\
\midrule
FSN, s0 & $\cdSfnZeroOne$ & $\cdSfnTwoThree$ & $\cdSfnFourSeven$ & $\cdSfnEightFifteen$ & $\cdSfnSixteen$ & $\cdSfnTwentyFour$ \\[-1pt]
 & \cic{-0.0072}{-0.0001} & \cic{-0.0144}{-0.0009} & \cic{-0.0905}{-0.0696} & \cic{-0.1525}{-0.1096} & \cic{-0.1114}{-0.0149} & \cic{-0.2049}{-0.1152} \\
FSN, s1 & $-0.0091$ & $+0.0001$ & $-0.0617$ & $-0.1074$ & $-0.0631$ & $-0.1505$ \\[-1pt]
 & \cic{-0.0128}{-0.0053} & \cic{-0.0062}{+0.0067} & \cic{-0.0719}{-0.0517} & \cic{-0.1305}{-0.0848} & \cic{-0.1142}{-0.0223} & \cic{-0.1975}{-0.1102} \\
FSN, s2 & $-0.0035$ & $-0.0023$ & $-0.0745$ & $-0.1294$ & $-0.1914$ & $-0.1198$ \\[-1pt]
 & \cic{-0.0073}{+0.0003} & \cic{-0.0086}{+0.0040} & \cic{-0.0852}{-0.0645} & \cic{-0.1526}{-0.1068} & \cic{-0.2357}{-0.1552} & \cic{-0.1591}{-0.0831} \\
FSN, mean & $\cdSfnZeroOneMean$ & $\cdSfnTwoThreeMean$ & $\cdSfnFourSevenMean$ & $\cdSfnEightFifteenMean$ & $\cdSfnSixteenMean$ & $\cdSfnTwentyFourMean$ \\[-1pt]
 & {\tiny range $\cdSfnZeroOneRange$} & {\tiny range $\cdSfnTwoThreeRange$} & {\tiny range $\cdSfnFourSevenRange$} & {\tiny range $\cdSfnEightFifteenRange$} & {\tiny range $\cdSfnSixteenRange$} & {\tiny range $\cdSfnTwentyFourRange$} \\
\midrule
No phases, s0 & $\cdNophaseZeroOne$ & $\cdNophaseTwoThree$ & $\cdNophaseFourSeven$ & $\cdNophaseEightFifteen$ & $\cdNophaseSixteen$ & $\cdNophaseTwentyFour$ \\[-1pt]
 & \cic{-0.0072}{+0.0007} & \cic{-0.0056}{+0.0068} & \cic{-0.0835}{-0.0620} & \cic{-0.1525}{-0.1072} & \cic{-0.1252}{-0.0452} & \cic{-0.1864}{-0.1110} \\
No phases, s1 & $-0.0004$ & $+0.0048$ & $-0.0692$ & $-0.1281$ & $-0.1828$ & $-0.1284$ \\[-1pt]
 & \cic{-0.0041}{+0.0033} & \cic{-0.0015}{+0.0111} & \cic{-0.0796}{-0.0591} & \cic{-0.1526}{-0.1051} & \cic{-0.2121}{-0.1582} & \cic{-0.1597}{-0.1005} \\
No phases, s2 & $+0.0022$ & $-0.0108$ & $-0.0798$ & $-0.1326$ & $-0.2258$ & $-0.1281$ \\[-1pt]
 & \cic{-0.0014}{+0.0059} & \cic{-0.0172}{-0.0042} & \cic{-0.0902}{-0.0697} & \cic{-0.1561}{-0.1093} & \cic{-0.2573}{-0.1985} & \cic{-0.1643}{-0.0975} \\
No phases, mean & $\cdNophaseZeroOneMean$ & $\cdNophaseTwoThreeMean$ & $\cdNophaseFourSevenMean$ & $\cdNophaseEightFifteenMean$ & $\cdNophaseSixteenMean$ & $\cdNophaseTwentyFourMean$ \\[-1pt]
 & {\tiny range $\cdNophaseZeroOneRange$} & {\tiny range $\cdNophaseTwoThreeRange$} & {\tiny range $\cdNophaseFourSevenRange$} & {\tiny range $\cdNophaseEightFifteenRange$} & {\tiny range $\cdNophaseSixteenRange$} & {\tiny range $\cdNophaseTwentyFourRange$} \\
\bottomrule
\end{tabular}
\end{table}

\paragraph{Reading the table.} Every base-layer seed is worse than the transformer in
every bin, and on every seed the dominant failure is the sixteen-to-twenty-three bin,
with per-seed deficits between $\cdBaseSixteenMin$ and $\cdBaseSixteenMax$ bits per
character around the seed mean of $\cdBaseSixteenMean$. Every FSN seed reverses the
deficit on every bin of depth four and beyond, and the window-cluster confidence interval
excludes zero in all twelve of those seed-bin cells. The FSN's shallow-bin margins are
marginal on every seed, and several of the shallow confidence intervals cross zero. The
no-phase configuration tracks the FSN on every deep bin on all three of its seeds, with
seed means of $\cdNophaseFourSevenMean$, $\cdNophaseEightFifteenMean$,
$\cdNophaseSixteenMean$, and $\cdNophaseTwentyFourMean$ on the four deep bins against the
FSN's $\cdSfnFourSevenMean$, $\cdSfnEightFifteenMean$, $\cdSfnSixteenMean$, and
$\cdSfnTwentyFourMean$, and its deep-bin confidence intervals likewise exclude zero on
every seed, so the deep-copy effect does not depend on the static frustration phases. The
widest seed spread in every family falls on the sixteen-to-twenty-three bin (ranges
$\cdBaseSixteenRange$, $\cdSfnSixteenRange$, and $\cdNophaseSixteenRange$ for the base
layer, the FSN, and the no-phase configuration respectively), which the per-depth curve
below explains. These margins are per-bin paired differences on the depth-binned token
subsets, with the deep bins pooled across the standard and enriched slices. They are not
whole-corpus validation values, and the whole-corpus comparison is Table~\ref{tab:main}.

\begin{figure}[t]
\centering
% Source: figures/fig_copydepth_curve.pdf built by figures/make_fig_copydepth.py from
% results/exploratory_figs/copydepth_curve.npz, written by
% experiments/scripts/exploratory/copydepth_stats.py (2026-06-12; full log
% results/exploratory_figs/copydepth_stats.out). Seed-mean margins over base
% comp_tan_1m_s0/s1/s2 and FSN sakanoise50_s0/s1/s2 vs comp_xf_1m_s0 ep48; bands are
% 95% position-bootstrap intervals, 2000 resamples.
\includegraphics[width=\linewidth]{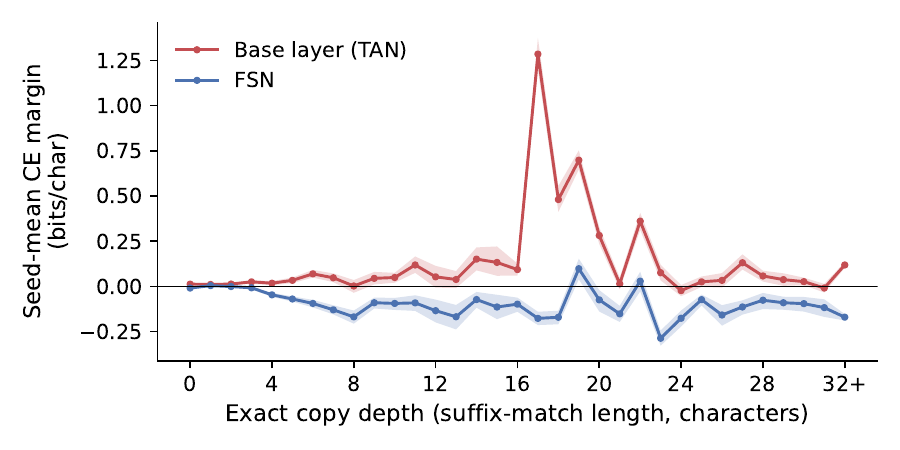}
\caption{Seed-mean cross-entropy margin against the converged transformer at every exact
copy depth, in bits per character, for the base layer and the FSN (three seeds each).
Shaded bands are ninety-five-percent position-bootstrap intervals around the seed mean
(two thousand resamples). The final point pools every position of depth thirty-two and
beyond. Depths below sixteen use the standard slice, and depths of sixteen and beyond
pool the standard and enriched slices, exactly as in Table~\ref{tab:copydepth}. The base
layer's deficit on the sixteen-to-twenty-three bin does not rise uniformly with depth.
It concentrates in spikes at depths seventeen, nineteen, and twenty-two, which
correspond to specific long repeated-content windows in the validation data.}
\label{fig:copydepthcurve}
\end{figure}

\paragraph{Margin as a function of exact depth.} Figure~\ref{fig:copydepthcurve} plots
the seed-mean margin of the base layer and of the FSN at every exact depth
$\ell = 0, \dots, 32$, where the final point pools every position of depth thirty-two
and beyond. The base layer's sixteen-to-twenty-three deficit is not a smooth function of
depth: it concentrates in spikes at specific depths, reaching $\cdCurveSeventeen$ bits per
character at depth seventeen, $\cdCurveNineteen$ at depth nineteen, and $\cdCurveTwentyTwo$
at depth twenty-two, while the base layer's margins at the nearby depths twenty-one,
twenty-four, and thirty-one are near zero. Positions sharing one exact depth in this range
come from a small number of long repeated-content windows in the validation data, so the
binned sixteen-to-twenty-three deficit is a severe failure on specific long-repetition
windows, replicated across all three base-layer seeds, rather than a uniform law of depth.
The same window structure explains the wide sixteen-to-twenty-three seed ranges in
Table~\ref{tab:copydepth}. The FSN's seed-mean curve is below zero at every depth of
four and beyond except depths nineteen and twenty-two, the two depths inside the base
layer's largest spikes, and at those two depths the FSN's positive margin is far smaller
than the base layer's deficit.

\paragraph{The code battery.} Table~\ref{tab:copydepthcode} reports the same measurement
on the code corpus, quoted in Section~\ref{sec:experiments}. The instrument is identical
to the enwik8 battery, with the same slice sizes, the same bins, and the same
window-cluster bootstrap, applied to the codeparrot validation split with a depth-label
cache computed from the code data. The code tier
has no fifty-epoch convergence arm, so the reference is the mean over the three
transformer code seeds of the per-token cross-entropy at each seed's best epoch rather
than a single converged run, and each margin is the paired difference against that
three-seed mean on identical tokens. The code corpus also has a larger share of deep-copy
positions, the regime where the FSN's advantage is largest, with \codeDeepShare{} percent of
code validation tokens at depth twenty-four or beyond against \enwikDeepShare{} percent
on enwik8. The evaluated checkpoints are the three FSN code seeds and the no-phase code
seed, each at its best validation epoch.

\begin{table}[t]
\centering
\caption{Per-seed copy-depth margins on the code corpus against the three-seed mean of
the transformer code baseline's per-token cross-entropy (each transformer seed at its
best epoch). Each cell is the paired per-token cross-entropy difference in bits per
character, model minus reference, on the bin's tokens, with the window-cluster bootstrap
ninety-five-percent confidence interval beneath it. Negative values favor the evaluated
model. Bins of depth sixteen and beyond pool the standard and the deep-copy-enriched
slices. Seed 1 of the FSN is the plateau-failure seed discussed in
Section~\ref{sec:experiments}.}
\label{tab:copydepthcode}
\scriptsize
\setlength{\tabcolsep}{2pt}
% Source: experiments/results/daido_probe/copydepth_code.json (corpus-generic battery,
% scripts/exploratory/copydepth_battery.py, 2026-06-12); margins recomputed in bits from
% the per-token CE cache results/exploratory_figs/copydepth_code_ce_cache.npz against the
% xfmcode30_s0/s1/s2 (best epochs 29/27/30) three-seed-mean reference, with the shallow
% bins rebinned 0-1/2-3 to match tab:copydepth; depth labels
% results/exploratory_figs/w2_copydepth_code_cache.npz. CIs are window-cluster bootstrap,
% 4,000 resamples, seed 0. Macro'd cells are the \cdc* macros of numbers.tex; the
% remaining literals are listed in the numbers.tex source comments.
\begin{tabular}{lrrrrrr}
\toprule
 & 0--1 & 2--3 & 4--7 & 8--15 & 16--23 & 24--32 \\
Tokens & \cdcNtokZeroOne & \cdcNtokTwoThree & \cdcNtokFourSeven & \cdcNtokEightFifteen & \cdcNtokSixteen & \cdcNtokTwentyFour \\
\midrule
FSN, s0 & $\cdcSfnZeroOne$ & $\cdcSfnTwoThree$ & $\cdcSfnFourSeven$ & $\cdcSfnEightFifteen$ & $\cdcSfnSixteen$ & $\cdcSfnTwentyFour$ \\[-1pt]
 & \cic{-0.0147}{-0.0066} & \cic{-0.0086}{+0.0027} & \cic{-0.0121}{-0.0034} & \cic{-0.0303}{-0.0199} & \cic{-0.0597}{-0.0356} & \cic{-0.0646}{-0.0398} \\
FSN, s1 & $\cdcFailZeroOne$ & $+0.0340$ & $+0.0005$ & $+0.0011$ & $\cdcFailSixteen$ & $\cdcFailTwentyFour$ \\[-1pt]
 & \cic{+0.2169}{+0.2313} & \cic{+0.0274}{+0.0404} & \cic{-0.0046}{+0.0056} & \cic{-0.0037}{+0.0060} & \cic{-0.0248}{-0.0086} & \cic{-0.0427}{-0.0185} \\
FSN, s2 & $-0.0071$ & $+0.0020$ & $-0.0048$ & $-0.0165$ & $-0.0338$ & $-0.0410$ \\[-1pt]
 & \cic{-0.0112}{-0.0030} & \cic{-0.0034}{+0.0074} & \cic{-0.0088}{-0.0007} & \cic{-0.0208}{-0.0121} & \cic{-0.0424}{-0.0264} & \cic{-0.0547}{-0.0304} \\
No phases, s0 & $\cdcNophaseZeroOne$ & $\cdcNophaseTwoThree$ & $\cdcNophaseFourSeven$ & $\cdcNophaseEightFifteen$ & $\cdcNophaseSixteen$ & $\cdcNophaseTwentyFour$ \\[-1pt]
 & \cic{-0.0260}{-0.0179} & \cic{-0.0040}{+0.0068} & \cic{-0.0042}{+0.0042} & \cic{-0.0187}{-0.0090} & \cic{-0.0387}{-0.0214} & \cic{-0.0453}{-0.0246} \\
\bottomrule
\end{tabular}
\end{table}

\paragraph{Reading the code table.} On the two healthy FSN seeds and on the no-phase
seed, the advantage over the reference grows with depth across the four bins of depth
four and beyond, and the confidence intervals on the two deepest bins are clear of zero
for every row of the table, including the plateau-failure seed. That failed seed is
behind the reference by $\cdcFailZeroOne$ bits per character on the shallowest bin and by
smaller amounts through depth fifteen, yet it holds deep-bin advantages of
$\cdcFailSixteen$ and $\cdcFailTwentyFour$, so the deep-copy capability remains present in a
seed whose general convergence failed. The code margins are smaller in absolute terms than the
enwik8 margins of Table~\ref{tab:copydepth}, partly because the reference here is an
epoch-matched seed mean rather than a converged single run, and partly because deep-copy
positions on code, while far more numerous, are individually less surprising under both
% architectures: deep-bin absolute CE (verified 2026-06-12 from the two CE caches, seed-0
% checkpoints, bits) is 0.148/0.156 (FSN) and 0.205/0.210 (XF) on code 16-23/24-32,
% against 0.341/0.189 (FSN) and 0.398/0.346 (converged XF) on enwik8.
architectures, with deep-bin absolute cross-entropies below their enwik8 counterparts in
every cell we checked.

\paragraph{Early checkpoint transfer.} The early-checkpoint result quoted in
Section~\ref{sec:capability} uses the same machinery with the earlier shallow binning
($0$ and $1$--$3$ in place of $0$--$1$ and $2$--$3$; the bins of depth four and beyond
are identical). It evaluates an FSN-family checkpoint trained for eleven epochs
(validation loss $\leanEpEleven$; the same stack as the FSN of Table~\ref{tab:config}
with the deterministic form of the kernel initialization, that is, the lean operating
point without the symmetry-breaking noise) against the same converged transformer
reference. The margins of $\copyFourSeven$ and $\copyEightFifteen$ bits per character on
the middle bins are measured on the standard slice. The $\copySixteen$ margin on the
sixteen-to-twenty-three bin pools the standard and enriched slices, with the two slices
individually giving $\copySixteenStd$ and $\copySixteenEnr$. The deepest bin, depth
twenty-four to thirty-two, pools the same two slices and gives $\copyTwentyFour$, with
the slices individually giving $\copyTwentyFourStd$ and $\copyTwentyFourEnr$. On the
shallow bins, where the transformer's additional epochs of general convergence dominate,
the eleven-epoch checkpoint remains behind the converged transformer.

\end{document}